\documentclass[journal]{IAENGtran}
\ifCLASSINFOpdf
   \usepackage[pdftex]{graphicx}
   \DeclareGraphicsExtensions{.pdf,.jpeg,.png}
\else
   \usepackage[dvips]{graphicx}
   \DeclareGraphicsExtensions{.eps}
\fi

\usepackage{mwe}
\usepackage[english]{babel}
\usepackage{t1enc}
\usepackage{graphicx}
\usepackage{epstopdf}
\usepackage[para]{footmisc}
\usepackage{setspace}
\usepackage{booktabs} 
\usepackage{longtable} 
\usepackage[english]{babel}
\usepackage[T1]{fontenc}
\usepackage[utf8]{inputenc}
\usepackage{wrapfig}
\usepackage{lscape}
\usepackage{rotating}
\usepackage{epstopdf}
\usepackage{siunitx}
\usepackage{paralist}
\usepackage{amsthm}
\theoremstyle{definition}
\usepackage{amsmath} 
\usepackage{enumerate} 
\usepackage{subcaption}
\usepackage{subcaption}
\usepackage{tabularx}
\usepackage{tabulary}
\usepackage{algorithm}
\usepackage{algorithmic}
{}
{}
{}

\begin{document}
%
\title{Cyber-Forensic Review of Human Footprint and Gait for Personal Identification}
%
%
%

\author{Kapil~Kumar~Nagwanshi,~\IAENGmembership{Member,~IAENG}
\thanks{Manuscript received April 13, 2018; Accepted May 23, 2019 .}
\thanks{K.K. Nagwanshi is with the Department
of Information Technology, MPSTME, Shirpur Campus, SVKM's Narsee Monjee Institute of Management Studies University Mumbai, India, e-mail: kapilkn@ieee.org.}
}

\maketitle

\pagestyle{empty}
\thispagestyle{empty}

\begin{abstract}
The human footprint is having a unique set of ridges unmatched by any other human being, and therefore it can be used in different identity documents for example birth certificate, Indian biometric identification system AADHAR card, driving license, PAN card, and passport. There are many instances of the crime scene where an accused must walk around and left the footwear impressions as well as barefoot prints and therefore, it is very crucial to recovering the footprints from identifying the criminals. Footprint-based biometric is a considerably newer technique for personal identification. Fingerprints, retina, iris and face recognition are the methods most useful for attendance record of the person. This time the world is facing the problem of global terrorism. It is challenging to identify the terrorist because they are living as regular as the citizens do. Their soft target includes the industries of special interests such as defence, silicon and nanotechnology chip manufacturing units, pharmacy sectors. They pretend themselves as religious persons, so temples and other holy places, even in markets is in their targets. These are the places where one can obtain their footprints quickly. The gait itself is sufficient to predict the behaviour of the suspects. The present research is driven to identify the usefulness of footprint and gait as an alternative to personal identification.
\end{abstract}

\begin{IAENGkeywords}
Biometric,  Gait, Human footprint, recognition,HMM, Access Control, Security.
\end{IAENGkeywords}

%
\IAENGpeerreviewmaketitle

\section{Introduction}\label{sec1}
The human brain recognizes and classifies objects, person, or places in an efficient, quick and smooth manner. For example, a person can identify his childhood friend comfortably in an old group photograph without encountering any difficulty. Since the recognition process happens so automatic and fast, therefore, it is difficult to translate this behaviour into a computer algorithm as perfect as the human being. In other words, it is neither permissible to define nor to observe meticulously whence the recognition process acts. Everything in this world, from the universe to an atom like oxygen is a  pattern.  Water is a  pattern of an assembly of hydrogen and oxygen; a human face is a pattern based on the contour of the eye, ear, hair, lips, cheek, etc. Likewise, the biometric oriented system is also a pattern based on biological matrices. The human footprint has an enormous legal capacity in Austria and some other countries such that one cannot ignore the role of biometric in society. Biometric passports are nowadays standard in India for travel across the world. Travelling by air is also very common; therefore, terrorists can take benefit of the vulnerable identification system available in railways, bus terminals or any such crowded spots to harm the assets of the nation. One must have a full-proof system which is capable of distinguishing people quickly to ensure the security of such stated places. Instead of the accomplishment of monetary transactions, the overpass of international borders and many more urgent solicitations, they are accountable for personal identification, which might be unavoidable. A traditional token-based method based on the passport, credit cards, etc. and knowledge-based identification using own identification numbers, passwords do not inherently imply the right capabilities. One may not be competent to at a guess that the person involved in the transaction is correct one, or on the other side, we are struggling with non-repudiation problem of information and personal security.

The biometric-based personal identification system is the traditional and most appropriate method for authentication if we compare it to smart card based or password-based methods\cite{Jain99, Jain2000,Franke2002,Jain04,Jain2004}. Fingerprint as a personal biometric proves it's capability as the signature from ancient time (around 1000 BC) used in different geographical locations Babylon pressed the tips of their finger into clay to record business transactions, China used ink-on-paper foot and finger impressions for business and to help identify their children, Nova Scotia, and Persia. On the epidermis portion of the body for example, finger, palm, the sole, and toe determine by different friction ridge lines present on it. These friction ridges make an impression on surfaces such as glassware, plastic articles, paper sheet, and wooden items. Ref.~\cite{Grew1684} gives a profound observation on friction ridge skin of the finger, palm, and foot. In the year 1686, Macello Malpighi observed loops, whorls, arches, and ridges in fingerprints as the unique pattern for identification. Ref.~\cite{Purk1823} documented nine distinct patterns to help identify types of prints. Regardless of this invention, the use of fingerprints did not catch on then.

In the year 1858 first legal contract between Sir William James Herschel, British Administrator of the Hooghly district in West Bengal, and Rajyadhar Konai, a local businessman, first used fingerprints on domestic procurement. In the year 1901 Edward Henry, an Inspector General of Police in Bengal, India, develops the first system of classifying fingerprints was first adopted as the official system in England, and eventually spread throughout the world\cite{Henry1900}. 

The primary function of biometric to verify and identify a genuine person or importer as tools of surveillance that scan for terrorist or unlawful activity for gaining access to a secret system. The automated system is necessary and conforms enhanced security standards despite technical limitations, but misuse of identity theft by criminals leads to hazards in the society.  Fingerprints, face, and eye are the common biometric traits used commonly for attendance, in ATMs to withdraw money, to purchase new mobile connection or in driving license \cite{Jain99,Bolle04a}. Interpol presentation authored by \cite{Farelo09}, presented a historical background of biometric traits and its utility in crime scene investigation from the age of non-electricity time. In addition, with fingerprints, face and retinal images, palm prints have been used in various security applications such as social security cards and passports \cite{Hen07,Ross03,You04,Zhang09,Simp10}. Hand shape and texture can also be helpful in personal recognition, but hand shape is not sufficient; hence another modality should be added to obtain the recognition pattern \cite{Kum06,Yoruk06}. 

Ref.~\cite{Rob78} had acutely examined for the individuality of human footprints, entire report on several physical characteristics of the individuals who made them was retrieved. The information on the footprint (and foot) morphology is especially significant because it explains the individuality of each person's footprints. The shape, or form, of an individual's foot, is unique his or her own should come as no surprise to physical anthropologists. The combined effects of heredity and life experiences are operative in determining the size and shape of our feet, and for each of us, the appearance of those consequences is uniquely our own.

According to Ref.~\cite{Vernon2009} {\bf{forensic podiatry}} is defined as the employment of in-depth and researched podiatry knowledge and expertise in forensic investigations, to confirm the relation of an individual with a  crime scene, or to solve any other legal interrogation concerned with the foot or footwear that requires knowledge of the functioning foot.  Forensic podiatrists support in the identification of perpetrators of crime where barefoot prints, footwear, and the CCTV gait proof are incorporated. The crime scene identification requires their expertise in the evaluation of the impressions of foot and lower limb use, the estimation, and matching of wear associated with the foot/shoe interface and in comparisons compelling evidence of shoe size. In their CCTV movie, forensic podiatrists examine the gait patterns of a person obtained on CCTV with suspected offenders. The identification of human remains from the comparison of the feet of the deceased with detail recorded in the podiatry reports of missing individuals requires forensic podiatrists. Ref.~\cite{Kanchan201429,krishan15} shows footprint for determination of sex based on BBAL--breadth of footprint at ball and BHEL--breadth of footprint at heel measurement. The HB Index of a footprint was derived as $(BHEL \div BBAL) \times 100$, for sex determination. It also shows different points for stature estimation from the anthropomorphic analysis of footprints\cite{Hairu15}.

Using footprint-based measurements constitutes one of many different possibilities to realize biometric authentication. Biometric amalgamation is a general method to cope with reduced recognition rates or performance of unimodal systems \cite{Malt09, Fathima14}. According to Ref.~\cite{Jain04}, most biometric systems incorporate multiple sensors, this effort functions on optical single-sensor output and may, therefore, assimilating manifold exemplifications and matching algorithms for the equivalent biometric classified as a blending or fusion method. The term recognition in proposed work refers to both verification (1:1 comparison with a claimed identity template) and identification (1: n comparison to find a matching template, if existing), as is used in several literature\cite{Jain2004, Bolle04a, Kapil12,Kapil14}.

\begin{figure}[!t]
\centering
\includegraphics[width=8.5cm]{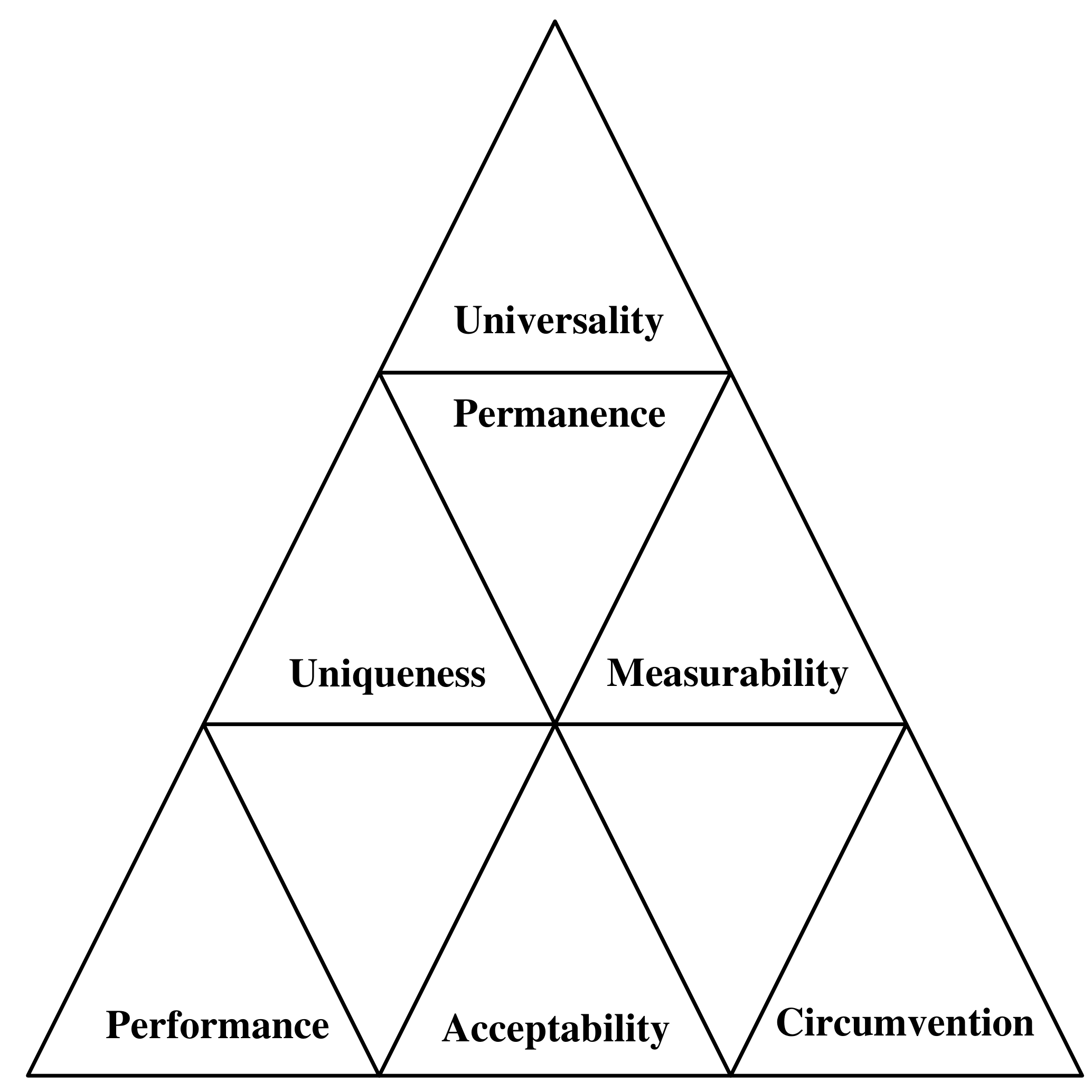}
\caption{Trade-offs between biometric systems\cite{Nagwanshi2017}. This figure has been created based on theoretical analysis of paper cited here.}
\label{fig1.5}
\end{figure}

Biometrics based personal identification and authentication has been used in different forms in our life \cite{Ross03,Bolle04a,Gonzalez06, Bhelkar17}. According to the National Science \& Technology Council (NSTC), \textbf{Biometrics} is defined as an application of statistical mathematics to biological sciences to recognize a person based on these statistical features or templates of fingerprints, palmprints, voice samples, retina, and footprints\cite{NSTC}. Ref.~\cite{Malt09} defined, \textbf{Verification} as a one-to-one association with a requested identity template and, \textbf{Identification} is defined as a process of one-to-many comparison to find a match. Figure~\ref{fig1.5} shows seven biometrics measures which have been discussed by Ref.~\cite{Jain99, Jain04,Nagwanshi2017} includes, 
\begin{inparaenum}[(i)]
\item	The \emph{universality}, is every subject in this system should hold the feature;
\item	these attained features must be effectively diverse for subjects in the applicable domain such that one can be distinguishable from another is \emph{uniqueness}; 
\item	the way in which a feature illustrates the discrepancy temporally is dealt with the permanence problem, and it is expected that the \emph{permanence} will be reasonably invariant over time for the specific biometric matching algorithm; 
\item	\emph{measurability} or \emph{collectability} is the simplicity of attainment or capacity of the feature that permits further processing and extraction of the applicable feature sets; 
\item	\emph{performance} is the measurement of accuracy, speed, and robustness of algorithm used; 
\item	\emph{acceptability} is the measurement of acceptance of the technology with respect to the biometric feature captured and arbitrated, and, 
\item	\emph{circumvention} denotes to the ease with which a trait might be counterfeit. No single biometric system from physiological to behavioural meet all the requirements of each promising application.
\end{inparaenum}

\section{Motivation}
Ref.~\cite{Nap04} presented an analysis of foot and footwear characteristics, impressions, and track ways that lead to significant evidence in a crime scene investigation.  Ref.~\cite{Uhl07,Uhl08,Wild08} have implemented a system based on eigenfeet, ballprint and foot geometry for the small dataset of 16 entries. Ref.~\cite{Nap04}, have published a criminological and analytical study on footprint and footwear. Ref.~\cite{Krishan08}, had applied footprint and foot outline dimensions for a particular tribe of criminal background, which helps in an estimation of the build of a criminal. His study also signifies the utmost importance of a footprint into understanding the crime scene investigations. Ref.~\cite{Wang09} give a model based on neural networks and fuzzy logic with a claimed recognition rate of 92.8\%. Ref.~\cite{Kumar10,Kumar11a,Kumar11b,Kumar13,Kumar12} introduce modified sequential haar energy transform technique, and PCA for recognition. Ref.~\cite{Siva15} has extracted fingerprint minutiae feature and PCA footprint features of the newborn on Raspberry Pi system. Ref.~\cite{Kho15} exhibit the use of PCA and Independent Component Analysis (ICA) for footprint recognition. Ref.~\cite{Hash16} has used ANN for the feature extraction and recognition with a recognition rate of 92\%. Primarily based on gap evaluation following are the motivations to propose a new approach to locating the uniqueness of human footprint.
\begin{inparaenum}[(a)]
\item Personal identification using gait is complicated as it is based on behavioral patterns. It could be used for behavioral analysis of a person, but not for personal recognition. In Ref.~\cite{Jung02} used highly correlated data (as all the images were collected in a day) that gives 100 \% recognition rate with only five subjects  by using shoe-type sensor could be a significant constraint in the view of users. Ref.~\cite{Wang04} and Ref.~\cite{Yun07} has also done experimentation for gait based recognition, but addition of more subject lead to degradation in performance. 
\item In Ref.~\cite{Naka00}, used Euclidean distance based personal identification among ten persons  with a recognition rate of 85\% using normalized static footprint where user ought to make a stand up stance each time. This was very basic method give direction to implement a more realistic approach for footprint-based personal identification. 
\item Personal identification based on footprint cannot be use as a method for attendance monitoring, or biometric-based transactions. On the other hand, security of public places for example hospital, airport, religion places, nanotechnology based hi-tech industries etc. is a challenging task. To overcome from this problem footprints are critical to use in legal capacity to capture as other biometric modalities such as fingerprints, palmprints, and retina images (refer~\cite{Fathima14}).  
\item The number of subjects taken by different footprint-based methods are limited to very few subjects. This motivates to add more number of subjects. 
\end{inparaenum}

A minimal number of subjects say ten exposed in most of the methods for footprint identification. The proposed plan will utilize the fuzzy logic based method for personal identification. Considerably more than 300 subjects with the temporal aspect will be taken into account to fill the gap identified in existing processes.

\section{Objective}
Primary objectives of the proposed work are: 
\begin{inparaenum}[(i)]
\item to relate the traditional biometric features to an innovative biometric human footprint; 
\item to meet all aspects of existing biometric characteristics to a new biometric modality human footprint;
\item to study the applicability of biometric footprint approach in industries, government and public domain;
\item to rectify the recognition rate(to reach the required accuracy of 0.1\% False Match Rate (FMR, the rate of falsely as genuine classified imposters) at 0.1\% False Non-Match Rate (FNMR, the rate of falsely as imposter classified genuine users) to a set of different matchers based on soft-computing for the identification task); 
\item to ascertain the employability of the footprint with respect to temporal, and somatic features; 
\item to improve the speed of recognition with beyond 200 subjects; 
\item to implement the prototype schemes for footprint biometric in order to evaluate system properties, including accuracy and performance;
\item to promotes it to install in legal capacities to identify the cops,
\item to prove the efficiency of algorithm with respect to temporal aspect (2 month gap) and total number of subjects, and
\item to evaluate system properties, including accuracy and performance, prototype systems for footprint biometric.
\end{inparaenum}

\section{Review of Systems}
Despite installed CCTV it is tough to identify a crime suspect who wears masks in face and gloves on hand, but he was not able to hide his footprint at crime scene walking pattern or gait behavior\cite{Birch13}. Analysis of recorded gait pattern compared with the gait pattern of a suspect and was convicted and sentenced to prison. Forensic podiatrists solved a lot number of cases based on video footage of criminals and the available footprints from the spot. In the year 1862 barefoot print of Jessie McLachlan convicts her in the crime scene of a woman's murder \cite{Hamilton2013}.In a murder case at Waushara County, Wisconsin, sock-clad footprints appeared on the motel room floor encompassing the slain body of Robert Kasun. After the detailed analysis of the footprint and its comparison with prime suspect Petersen convicted a life sentence without parole\cite{Nire17}. Now it is very standard practice to check for foot related pieces of evidence, and it proposed the necessity of legal podiatry organization. Robert Kennedy of the Royal Canadian Mounted Police (RCMP) researched the uniqueness of footprints.  Started in the year 1990s with a collection of 24,000 footprints. The collection proves the hypothesis of the unique character of footprints. From a rigorous statistical analysis of the footprint data, Kennedy has concluded the chances of matching the footprint is one in 1.27 billion\cite{Nire89,Nire16,Nire17}. The American Society of Forensic Podiatry formed in the year 2003 by founder John DiMaggio with fifty-one members from different universities and research organizations. Most of the available literature concentrate on crime scene investigation\cite{Dim11}.  The forensic podiatry discipline focused on three main application areas:
\begin{inparaenum}[(a)]
\item Footprints (This is principle area of discussion in present work),
\item Gait Analysis based on CCTV footage, and
\item Footwear.
\end{inparaenum}
The previous section discusses the background and historical development of biometric system such as fingerprints, palmprints, and finally footprints. The present section gives a comprehensive literature survey on gait based recognition system, footprint-based recognition system, fuzzy logic, and analytical tools BigML{\textsuperscript{\textregistered}} and IBM{\textsuperscript{\textregistered}} Watson Analytics. 
 
\section{Gait-based system}
At the time of Aristotle (382–322 BCE), some evidence proves the applicability of gait examination. One exciting thing about India is the gait analysis of any person to see his/her characteristics for suitability of marriages. Giovanni Borelli (1608–1679) uses the first scientific gait analysis research \cite{Baker07}.
\subsection{Body Shape and Gait}
According to Ref.~\cite{Collins02}, a gait is determined using a person's four main walking stances as \begin{inparaenum}[i).] \item right double support -- both legs touch the ground and right leg in front, \item right midstance -- legs are closest together, and right leg touches the ground, \item left double support, and \item left midstance. \end{inparaenum} 
They use viewpoint based template matching of body silhouettes captures height, width, and part proportions of the human body(silhouette extraction), on the other side cyclic gait capture stride length and the amount of arm swing analysis to extract keyframes (identify key frames) to compare with the training frames (template matching). The frame matching gives statistical parameters and score matching (nearest neighbor classification using combined scores). Two templates $a$ and $b$ gives the matching score as eq.~\ref{eq1}
\begin{equation}
C(a,b)=\frac{max(\hat{a} \times \hat{b})}{max(\hat{a} \times \hat{a})max(\hat{b} \times \hat{b})} \label{eq1}
\end{equation}
where $\hat{x}$ is given by eq.~\ref{eq2} 
\begin{equation}
\hat{x}=\frac{(x-mean(x))}{\sigma(x)} \label{eq2}
\end{equation} 
and $\times$ is cross correlation operation. The maximum value of $C(a,b)$ is chosen. They used datasets from the MIT database shows 25 subjects (14 male, 11 female), the CMU MoBo database motion sequences of 25 subjects (23 male, two female) exercising on a treadmill, the U.Maryland database with 55 individuals (46 male, nine female), and the University of Southampton database contains 28 subjects walking indoors on track. Match scores for the across-gait condition vary from 76\% to 100\% on different profile test conditions. Recently, Zhang \textit{et al.} ~\cite{Zhang18} have developed a system for skin color detection based on segmentation of captured hand posture in a complex background. Based on findings, this method can be installed with gait and footprint-based system to recognize the behavior of a person. 
 
\subsection{Invariant Gait Recognition}
Ref.~\cite{Kale03} presents an invariant gait recognition approach. Human gait is spatiotemporal in nature. This algorithm gives a comprehensive solution to synthesizing arbitrary views of gait and applying incorporated views for gait recognition at any arbitrary angle to the camera on a perspective plane. Their dataset consists of 12 subjects, walking onward straight lines at varying values of azimuth angle $\theta$ from $\SI{0}{\degree}, \SI{15}{\degree}, \SI{30}{\degree}, \SI{45}{\degree}$  and to $\SI{60}{\degree}$ ($G_0,\ G_{15},\ G_{30}, \ G_{45}, \ and \ G_{60}$).   Initially, it starts with capturing the original video for estimating the azimuth angle; subsequently, the video sequence synthesized in a new view. A simple and accurate camera calibration procedure was also given. This algorithm uses very less number of subjects which is not sufficient to use in real life for person recognition. The recognition rate, FMR, FNMR, FRR, and so forth has not quoted in the literature.  
\subsection{Hidden Markove Model in gait}
Ref.~\cite{Boul05} shows the advanced research on identification by observation of a person's walking or running style(gait) and its challenges to prove that such a system is realistic and is likely to be developed and used in the years to come. Unobtrusiveness is the feature in which the prior consent of the subject is not required to capture the gait. The clothes, walking surface, walking speed, and emotions affect the gait pattern. Pattern ambiguity is a possibility in several cases; for example, the same kind of person (considering feelings) may have the same gait pattern. Adding another modality like the foot, retina, or palmprint (obtrusive) in conjunction with gait, can make this identifiable. But if we add pattern, it will need personal consent while foot pressure (requires special instruments to capture) or the face can help significantly. The calculation of two frequency templates (Fourier examination determines the harmonic component) is straightforward. On the other hand, in spatial template matching the sequence of gait features must be compared with another series. If $T1$ and $T2$ are the fundamental walking periods, then the cumulative distance is calculated by eq.~\ref{eq3}:
\begin{equation}
D_{12}=\frac{1}{U}\sum^{T}_{t=1}u(t)D(f_1(w_1(t )), f_2(w_2(t))) \label{eq3}
\end{equation}
where $w_1(t) $ and $w_2(t)$ are the wrapping functions, and $U$ is given by eq.~\ref{eq4} 
\begin{equation}
U=\sum^{T}_{t=1}u(t)\label{eq4}
\end{equation}
and $D(\cdot)$ gives the distance between feature vectors at time $t$. HMM is applied in gait stances. For a model $\lambda$, and test feature vectors $\tilde{f_i}$, the higher probability is acknowledged to be identical to the test subject of one of the models associated with the database with $N$ number of subjects sequences is given by eq.~\ref{eq5}

\begin{equation}
identity(i) =  \underset{j}{\operatorname{argmax}} \ P\left(\frac{\tilde{f_i}}{\lambda_{j}}\right),\  \forall j=1,2,..,N  \label{eq5}
\end{equation}

They manage 71 subjects in their database. The right item gives high confidence in top 10 matches. Gait alone is not sufficient to provide reliable and accurate recognition; this carries clear implication to use gait in conjunction with other biometrics traits for best performance in personal identification. Many researchers have adopted the gait-based method. But gait based technique is not full proof; hence it is needed to add some other biometric trait such as the face. In Japan and France, health-care personals are using gait to diagnose the problems that can be modelled with walking or fall pattern of a person Ref.~\cite{Kura05,Hew07}.  Walking pattern and stepping pattern is also known as dynamic footprint has been used to characterized biometric user identification. The principal component analysis is applied to identify the user with extracted features \cite{Yun07}. Ref.~\cite{Take09,Take10}, has employed a load distribution sensor based on fuzzy distribution to capture foot pressure from 30 volunteers with an estimated recognition rate of 86\%. Pressure sensor based system using a neural network has observed 28.6\% false acceptance rate \cite{Ye09}. Tivive et al.\cite{Tiv10} presented a technique based on radar Doppler spectrograms for gait classification denoted in the time-frequency domain. Sensor quality is having a high impact on the biometric-based system for recognition as it should be measurable, distinctive, and permanent over time \cite{Prabha11}. Ref.~\cite{Pata11},developed a gait based recognition system and obtained unique dynamic plantar pressure patterns among 104 individuals.  

\subsection{View Transformation Model in the Frequency Domain}
Ref.~\cite{Makihara2006} uses a total of 719 gait sequences from 20 subjects for gait recognition from 24 perspective directions at every 15 degrees using a view transformation model (VTM) in the frequency domain features to generate the training set of multiple subjects from features of one or a few of the view directions. Initially, the gait of a person is extracted from IR camera using temperature based background removal to construct an amplitude spectra for spatiotemporal gait silhouette volume(GSV) of a person scaling and registering the silhouette images. Gait periodicity extracts the frequency-domain features of the volume by Fourier analysis to generate training and test set for recognition. The verification rate at a false positive rate is 10 \% in the ROC (Receiver Operating Characteristics) curves and the averaged verification rate. These systems achieve higher performance than the earlier perspective projection (PP) system proposed by Ref.~\cite{Kale03}, except for the frontal view $G_0$.  
\subsection{Challenges in Gait Recognition}
Ref.~\cite{Nixon06} presented a summary of comparison of future challenges related to the gate. The typical assumption includes: 
\begin{inparaenum}[(i)] 
\item no camera motion, 
\item only one person on the view field, 
\item no occlusion, 
\item no carried objects,
\item normal walking motion,
\item moving on a flat ground plane,
\item constrained walking path,
\item plain background or environment,
\item specific viewing angle, and 
\item data recorded over one time-span.
\end{inparaenum}
 
While, the suggested really possible affecting factors includes: 
\begin{inparaenum}[(i)] 
\item changes to the clothing style, 
\item distance between the camera and the walker, 
\item background or environment, 
\item carried objects such as a briefcase, 
\item abnormal walking style, 
\item variations in the camera viewing angles, 
\item walking surfaces such as playground, grass or slope,
\item Mood,
\item walking speed, and 
\item change with time.
\end{inparaenum}

\subsection{Gait for Fall Prevention}
Ref.~\cite{Hew07} develops a technique of gait detection and a balanced signature extracted from a force plate to fall prevention in old aged person in the PARAchute project. Every year in France, more than 9000 death tolls due to the falling of the senior citizens, which is the most prominent medical challenge. Their system is quite simple and does not require the third person to monitor the elderly without intervening daily life; hence, privacy is preserved. It can function independently as a part of the home security network. The gait is analyzed based on video camera sequencing, while the balance is assessed using a small force plate.  The force plate measures centre of pressure (COP) displacement in anteroposterior (AP), mediolateral (ML), and resultant (RD) directions to describing the static and dynamic stability with no limitation on the posture of the person. The obtained gait sequence enables the calculation of kinematic and spatiotemporal variables already identified as appropriate in an examination of the risk of falling. For better results, this system requires large dataset as well as data fusion of home-based system and the clinical data.  

\subsection{Floor Pressure Sensing and Analysis}
Ref.~\cite{Qian2008,Qian2010} employs 3D trajectories of the centre of foot pressures atop a footstep include both the single dimensional pressure silhouette and 2D position trajectories of the COP taken by a large area high-resolution pressure-sensing floor. A set of features has extracted based on the 3D COP trajectories can be further used for personal identification together with other features such as stride length and cadence. The Fisher linear discriminant classifies the foot pressure. This approach has tested for a floor pressure dataset collected from 11 subjects in different walking styles, including varying speed walking and freestyle walking yields identification results with an average recognition rate of 92.3\% and FAR of 6.79\%.
\subsection{Radar Doppler Spectrograms Gait Classification}
Ref.~\cite{Tiv10}introduces a radar Doppler image classification approach, for human gait based following the time-frequency domain from five subjects moving towards and away from the radar by \SI{0}{\degree} and \SI{30}{\degree} aspect angle including the nonobstructed line of sight. The initial two stages of this method are intended to extract Doppler features that can adequately describe human motion based on the nature of arm swings as free-arm swings, one-arm confined swings, and no-arm swings, and the third stage performs multi-layer perceptron (MLP) classification. The one-arm confined swings and no-arm swings can depict that individual carrying objects or, he is in stressed conditions. Maximum classification rate obtained was 91.2 \% in two arm motion, but the number of subjects taken was very less for a practical purpose. 
\subsection{Clothing-invariant Gait Identification}
Ref.~\cite{Hossain10} uses part-based clothing categorization and adaptive weight control for the identification of gait. It is tough to identify a gait while variations in clothing alter an individual's appearance due to change of the pattern of silhouettes of certain parts hence a part-based procedure has the implied on selecting the relevant sections of a gait. The clothing used in the data set of 48 subjects both male and female are Regular pants, Half shirt, Casual wear, Baggy Pants Full shirt, Raincoat, Short Pants Long Coat, Hat, Skirt, Parker, Casquette Cap, Casual Pants, Down Jacket and Muffler. The human body has divided into eight sections viz neck, waist, knee, pelvis, waist-knee, neck-knee, neck-pelvis, and pelvis-knee with four overlapping and the more substantial parts have a higher discrimination capability, whereas the smaller pieces are also seeming to be unaffected by clothing alterations. Some clothes are common to different elements can be classified in a collective group. This technique exploits the discrimination capability using an equal weight for each part and controls the weights adaptively meant on the distribution of distances between the test set with the data set. The EER varies from 0 to 0.5\% depending on clothing combination. 

\subsection{Geometric Features and Soft Biometrics based Gait Recognition}
Ref.~\cite{Moustakas10}introduced a new framework for gait recognition expanded with soft biometric information acquired as height and stride length information has employed in a stochastic framework for the detection of soft-biometric features of high discrimination ability. The captured gait sequences are processed to extract the geometric gait features based on Radon transforms and on gait energy images and the soft biometric features. Ref.~\cite{Ioannidis07} utilized the calibrated stereoscopic sequences based on real-world coordinates and absolute distances captured in HUMABIO and ACTIBIO databases for the estimation of height and the stride length soft biometric features. These sequences trivially overcome in the determination of the features that resemble the highest-lowest part of the subject, concerning height, and to the most significant distance between the legs in a gait cycle. Different characteristic plots in the paper demonstrate the efficiency of their approach, applying various algorithms for gait feature extraction and different databases. Four curves have shown on each graph corresponding to the cumulative matching scores (CMS) using the gait feature (gait) only or coupled with height (gait+height), or stride length (gait+stride), or both soft biometrics (gait+height+stride). On extending the gait feature with added soft-biometric information significantly enhances the gait recognition efficiency. This method reduces the FAR and FRR in the EER point from 4\% to 15\% decrement in the ACTIBIO database depending on the soft biometrics applied.

\subsection{UDSA for View-invariant Gait Signal Computing}
Ref.~\cite{Lu2010} proposes a supervised manifold learning technique, known as uncorrelated discriminant simplex analysis (UDSA), for view-invariant gait signal computing. They have selected gait sequences of 124 subjects obtained from two distinct views, i.e., \SI{36}{\degree} and \SI{90}{\degree}. This paper explains the significance of human gait as a valuable biometric signature which has adopted as a fundamental domain for computer vision researchers. Gait has taken due to its applicability in the automatic human identification and gender recognition from a distance in several surveillance applications. The process begins with the collection of human gait sequences from different views, one for training and one for testing. The primary goal of extracting the gait features in such a way that the relationships among intraclass samples are small and those of interclass samples are high has achieved by the statistical learning framework. Gait has recognized by two different subjects, and gender has identified, by the male and female classes. However, two items are not sufficient to determine the majority of recognition problems. To address this issue, they revised their approach by a feature subspace to unfold the interclass samples obtained from the same view expected to be separated, and remove the intraclass samples obtained from different perspectives to be clustered together to enhance the discriminant power for recognition. After collecting gait sequences, the Gait Energy Image (GEI) features for intraclass gait sequences have extracted because it preserves the dynamic and static information into a gait sequence from two different views and takes their difference to characterize the within-class information. Next step is to derive a linear subspace from different learning received to predict the intraclass gait sequences collected from divergent views to be as close as possible and the interclass sequences obtained from the same perspective be as far apart as possible. The major weakness of UDSA approach is the view of the testing gait sequence is considered to be known before recognition, which means that there should be one view-estimation module before view-invariant gait signal computing. The recognition accuracies using the UDSA method varies from 50\% to 90\% depending upon the view angle.

\section{Footprint-biometric recognition}
This section covers a detailed literature survey of a footprint-based biometric technique. Some of the pieces of literature are intended to use the pressure sensing technique while others are based on optical sensing and imaging technique. 
\subsection{The BIG-MAT system}
Ref.~\cite{Naka00} used the BIG-MAT on the hard floor of size $440 \times 480\  mm^2$ with the load cells forming the matrix of $ 10 \times 10 \ mm^2 \ (44 \times 48 \ cells)$ to acquired total 110 samples from ten volunteers of the pressure distribution of footprints. An image of 256 grey levels at a sampling speed of 30 fps for 5 sec acquire the barefoot prints.  The monochrome image is given by eq. \ref{eq6}: 
\begin{equation}
f(x,y):-X/2<x\leq X/2, -Y/2<y\leq Y/2 \label{eq6}
\end{equation}
represents the pressure distribution of a footprint. (where $X$ and $Y$ are widths of the mat). Initially, in the pre-processing phase, they use the raw footprint to translation motion of the whole image to the centre of mass then go for segmentation of in the form of separation of both feet. It is followed by calculation of the centre of the foot for both the feet. Then normalize the image on the centre of mass. Finally, reconstructed whole footprint image. In the recognition phase, they use the distance $\tau$  between the centre of mass of both footprints and the angle $\theta$  between left and right footprints before normalization for personal identification. Product of the first eigenvector of the left $e_L$ and right $e_R$ footprints calculates the angle using eq.~\ref{eq7}.
\begin{equation}
\theta=cos^{-1}({(e_L\cdot e_R)}/{(\parallel e_L \parallel \parallel e_R\parallel)}) \label{eq7}
\end{equation}
Employment of geometric elements $\theta$ and $\tau$ for recognition gives evaluation function as shown in eq. \ref{eq8}. 
\begin{equation}
\begin{split}
d =\sqrt{{\sum}_{(x,y)}\{I_p(x,y)-I_{{\tau}_n}(x,y)\}^2}\\
+\alpha\mid{\tau}_p-{\tau}_{r_n}\mid +\beta\mid {\theta}_p-{\theta}_{r_n}\mid \label{eq8}
\end{split}
\end{equation}
where $p$ is input footprint and $r_n$ is the $n^{th}$ registered footprint $(\forall n \mid n=1,2,..,N)$, and $N$ be the total number of subjects. The first term is the Euclidean distance between $I_p(x,y) $ and $I_{{\tau}_n}(x,y)$. The second and third terms are the absolute difference of the distance $\tau$ and the angle $\theta$ between the registered and input footprint before normalization. $\alpha$ and $\beta$ are weighting parameters. After a comparison of the input to all registered footprint, the minimum $d$ recognize entered footprint as the registered footprint. 85.00\% is the highest recognition rate achieved, which appears to be inadequate for practical uses. The author intends to apply personal identification for automatic data acquisition for people when many family members live together.
\subsection{The Pressure Mat using HMM}
Ref.~\cite{Jung02, Jung03a, Jung03b}, proposed an unconstrained person identification technique using dynamic footprint. They use one-step walking data through a mat-type pressure sensor to acquire quantized COP (Center of Pressure) trajectory from eight subjects. Like \cite {Naka00}, they calculate the width $w_x$ and $w_y$ of the sensor, in the same way by using eq.~\ref{eq9} (used $w_x$ for $X$ and $w_y$ for $Y$).  
\begin{equation}
f(x,y):-w_x/2<x\leq w_x/2, -w_y/2<y\leq w_y/2 \label{eq9}
\end{equation}
Similarly, they calculated eigenvector of the left $e_L$ and right $e_R$ footprints then create a covariance matrix $H$.  Levenberg-Marquart learning method combined with HMMs (Hidden Markov Model) for both the feet to achieve better performance by Levenberg-Marquart learning method. LM method is a blending of Newton method and gradient descent algorithm.Weight is updated by eq.~\ref{eq10}. 
\begin{equation}
\begin{split}
w(n+1)=w(n)-(Z^T \cdot Z+\lambda I)^{-1}\cdot Z^T \varepsilon(w(n))\\
(Z)_{ni}\equiv \frac{\partial \varepsilon^n}{ \partial w_i} \label{eq10}
\end{split}
\end{equation}
where $\varepsilon^n$ is the error for nth pattern,and $\lambda$ is step size. The recognition rate achieved by HMM blended Levenberg-Marquart learning method is only 64\% with a maximum of 80\% correctly classified samples of 11 subjects.

\subsection{Footprint Heavy Pressure Surface}
Ref.~\cite{Wang04}, uses Gauss curvature and average curvature on footprint heavy pressure surfaces pick-up to obtain the segmented footprints. The human physical characteristics decide heavy pressure surfaces to realize the identity through abstracting the foot-body pressure images and analyzing their shape features. In the preprocessing phase, Gauss function in eq.~\ref{eq11}: 
\begin{equation}
h(n)=c \cdot e^{{-n^2}/{2 \sigma^2} } \label{eq11}
\end{equation}
filters the image noise. Here normalization coefficient $c$ does not affect the mean of the original image. The samples ranges between $-3\sigma \leq n \leq 3\sigma$  used to satisfy eq.~\ref{eq12}.
\begin{equation}
{\sum}^{-\infty}_{\infty} h(n)=1 \label{eq12}
\end{equation}
where, $n$ be the slope of integer lies between $-N$ and $N$. Consider $f (u, v)$ be the distance from the dispersed curved surface $u$ to the parametric surface point $v$ in 3D space. Then eq.~\ref{eq13} represents a dispersible parametric surface.
\begin{equation}
\begin{split}
X(u,v)= [u, v, f (u, v)]^T\\ 
\forall\  u=1,2,..,m\ \\ 
\forall\  v=1,2,..,n \label{eq13}
\end{split}
\end{equation}
If $f_u$ and $f_v$ be the first order derivative and $f_{uu}$,$f_{vv}$, and $f_{uv}$ be the second order derivative of $f (u,v)$ then Gauss curvature is obtained as eq. \ref{eq14}, and average curvature is calculated as eq.~\ref{eq15}
\begin{equation}
H=\frac{(1+f^2_v)f_{uu}+(1+f^2_u)f_{vv}-2f_u f_v f_{uv}}{2(1+f^2_u+f^2_v)^{3/2}} \label{eq14} 
\end{equation}
\begin{equation}
K=\frac{f_{uu}f_{vv}-f^2_{uv}}{(1+f^2_u+f^2_v)^2} \label{eq15} 
\end{equation}
The Gauss curvature and average curvature represents the principal curvatures of the surface by eq.~\ref{eq16}
\begin{equation} \label{eq16}
\begin{split}
k_1=H+\sqrt{H^2-K}\\
k_2=H-\sqrt{H^2-K} \\
\forall \ k_1\geq k_2
\end{split}
\end{equation}
The pressure surface resembles spherical cap in geometry and has a vital role in the human identification. The geometrical shape index presented in eq.~\ref{eq17} gives an unique value for each footprint.
\begin{equation} \label{eq17}
S_i=\frac{2}{\pi} \times arctan \left(\frac{k_1+k_2} {k_1-k_2}\right), \ \forall S_1 \ \epsilon\  [-1,1]
\end{equation}
With 165 samples from 70 subjects, they show the simplicity and robustness of the method and broad application in the criminal investigation, but nothing has been written for matching score.
\subsection{Footprint Similarity for Gait Analysis}
Ref.~\cite{Kura05} has introduced a technique to estimate footprint comparison of the patient's walking pattern(gait) which is initially unstable, and as the treatment goes on it becomes stable to determine the health status. The method initially starts with the edge detection and noise reduction (using the median filter to reduce salt and pepper noise) of 1200 dpi footprint image. Laplace transform has employed for binarization of the footprint. Edge of the binary image used erosion and dilation for smoothening. The distance is given by $ d(1/m\times np)$  for all $np$ points from the centroid of a footprint. Eq. \ref{eq18} obtains the similarity measure it uses DFT of foot image
\begin{equation}\label{eq18}
d(t)\ =\ a_0+\sum^{m/2}_{n=1}(a_n\ cos\ 2n\pi t + b_n\ sin\ 2n \pi t)
\end{equation}
where $m$ is the even number of sampling points, $a_0, a_n,\ and\  b_n $ are the Fourier coefficients, $t$ is the normalization parameter ranges from $1/m \ to \ 1$. The spectrum is calculated as eq.~\ref{eq19}, likewise patient's spectrum is calculated by eq.~\ref{eq20}.
\begin{equation}\label{eq19}
c_n=\ \sqrt{a^2_n + b^2_n}\ ; \ \forall\  n = 1,2,3,..,m/2
\end{equation}
\begin{equation}\label{eq20}
\hat{c}_n\ =\  1,2,3,..,m/2
\end{equation}
The matching spectra between reference and matching footprint presented in eq.~\ref{eq21} gives the similarity of the footprint patterns.
\begin{equation}\label{eq21}
R=1\ - \frac{1}{2} \ \sum^{m/2}_{n=0}\ \mid c_n -\ \hat{c}_n \mid
\end{equation}
Inverse DFT removes high-frequency components to produce smoothed edges. The footprint is reconstructed using eq.~\ref{eq22}
\begin{equation}\label{eq22}
d(t)\ =\ a_0+\sum^{k}_{n=1}(a_n\ cos\ 2n\pi t + b_n\ sin\ 2n \pi t)
\end{equation}
If $k$ was the same number as in the previous step of Fourier transform, the reconstructed footprint pattern is the same as the original one. This method of footprint similarity has not used for personal identification preferably it is used to check the rehabilitation status of any person using the footprint and walking pattern.  

\subsection{Principal Component Analysis in Dynamic Footprint}
Ref.~\cite{Yun07} presented an approach for user identification using a dynamic footprint. While walking over the UbiFloorII consists of a $12 \times 2$ array of wooden tiles, it collects (walking pattern) gait samples and stepping pattern (dynamic footprints) samples from 10 subjects. Each wooden tile of size $30 \ cm^2$ each equipped with uniformly arranged 64 photo-interrupter sensors connected to a microcontroller for data acquisition through   CAN (Controller Area Network) cable. They create the $8 \times 4$ footprint model to include all possible footprints based on the centre of pressure applied by foot on walking. There are $2^{32}$ possibilities of a footprint vector using this array, but the ANN vector like $[-1,-1,-1,...,-1]$ and $[1,1,1,...,1]$ are not possible. Similarly, in some sensor dots of this array never touched by a foot due to its geometry.  The PCA and ANN module identifies the matched features. Result gives a recognition rate of 90\% to a maximum of 96\%. They also claim the viability of stepping pattern in the homelike environment as an encouraging means of the automated and transparent user identification system.
\subsection{Eigenfeet, Ballprint and Foot geometry biometrics}
\label{uhl-para}
Ref.~\cite{Uhl07,Uhl08}  introduced a method based on Eigenfeet, ballprint and foot geometry biometrics for personal identification. Three features (i). \emph{Eigenfeet} features for both shape and texture features like Eigenfaces used in face recognition described by \cite{Turk91}, (ii). \emph{Minutiae-based} ballprint features similar to the one used in fingerprint identification and verification systems \cite{Jain2004, Jain07, Malt09}, and (iii). \emph{Geometrical or statistical} features focusing on characteristics such as the foot widths, to improve recognition rates. Due to the practice of wearing shoes in regular activities, the footprint-based recognition system applies to the clean and comfortable areas where high security is not an issue with less privacy concern. Such application areas include public baths, spas and also Japanese apartments, unpaid admission to fee paid areas. The three modules of this system include: (i). \emph{Image acquisition} achieved by simple flatbed scanners at 600 dpi. (ii). \emph{Preprocessing} to normalize the image as per requirement and extract the distinct features of the footprint image. Canny edge detection with a threshold applied to original footprint image $B$ gives a binarized image $B_1$. 
\begin{equation}\label{eq23}
B_2(x,y)  =  max(bin_b(B)(x,y),B_1(x,y))
\end{equation}
$B_2(x,y)$ in eq.~\ref{eq23} is an image operated to a binary threshold value $b$ using operator $bin$ on $B$. This binarized image $B_2$ is reduced to morphological expansion using a square structuring component $S$ to meet the boundary using eq.~\ref{eq24}.
\begin{equation}\label{eq24}
B_3=B_2\ \oplus\ S =\ \{(x,y)\ | \ S_{x,y} \cap\ B_2 \neq\ \emptyset \ \}
\end{equation}
where $S_{x,y}$ is a shifting of $S$ by coordinate value of $(x,y)$. The binarized image $B_4$ is obtained using $B_3$ on the removal of small white Binary Large Objects (BLOBs) and a filling of whole black BLOBs except the background. Morphological erosion operation on $B_4$ is applied to get $B_5$ in  eq.~\ref{eq25}:
\begin{equation}\label{eq25}
B_5 = \ B_4\  \otimes\ S \ =\ \{ (x,y)\ | \ S_{x,y} \subseteq\  B_4 \}
\end{equation}
The angle between the $y-$ axis and the major axis $\Theta\ $  (calculated by eq.~\ref{eq27}-\ref{eq29}) estimates the best-fitting ellipse for rotational alignment of the matching footprint. Let $B$ is the binary $n\times m\ $ image with $A$ number of white pixels, then the centre of mass $C(\overline{x},\overline{y})\ $  is calculated by eq.~\ref{eq26} (for both ellipse and the binary image $B$). 
\begin{equation}\label{eq26}
\begin{split}
\overline{x}=\ \frac{1}{A}\ \sum^{n}_{i=1}\ \sum^{m}_{j=1}\ jB(i,j)\\
\overline{y}=\ \frac{1}{A}\ \sum^{n}_{i=1}\ \sum^{m}_{j=1}\ iB(i,j)
\end{split}
\end{equation}
Let,
\begin{equation}\label{eq27}
x'=x-\overline{x} \text{  and  } y'=y-\overline{y}
\end{equation}
then,
\begin{equation}\label{eq28}
\begin{split}
\Theta=\frac{1}{2}\ arctan\ \frac{2\  \mu_{1,1}}{ \mu_{2,0} -\mu_{0,2}}\\
\end{split}
\end{equation}
where,
\begin{equation}\label{eq29}
\begin{split}
\mu_{1,1}=\sum^{n}_{i=1}\ \sum^{m}_{j=1}\ (x'_{ij})^2\ B(i,j);\\
\mu_{2,0}=\sum^{n}_{i=1}\ \sum^{m}_{j=1}\ x'_{ij}\ y'_{ij}\ B(i,j)\\
\mu_{0,2}=\sum^{n}_{i=1}\ \sum^{m}_{j=1}\ (y'_{ij})^2\ B(i,j) 
\end{split}
\end{equation}
Three main features namely Eigenfeet using PCA-based approach, minutiae, and shape are utilized in recognition of footprint.
and (iii). \emph{Matching and Decision module} observed a value of FAR as 2.5\%, and FRR as 3.13\%. 

\subsection{Feature Selection and Matching}
Ref.~\cite{Wild08} elaborated different features for footprint matching.  The \emph{silhouette} is a feature based on contour distance in two dimensions onto the eigenspace spanned by the 40, 100, 200 or 400 most significant eigenvectors of covariance matrix $C$ to the centroid, length and enclosed area of silhouette polygon it requires \emph{dynamic time warp (DTW)} matching classifier \cite{Myers81}.  Let $s_k$ be the contour distances given by eq.~\ref{eq30}. 
\begin{equation}\label{eq30}
s_k\ =\ |S_k\ -\ C|\ \forall k \in \{1,..,l(S)\}
\end{equation}
where $S\ =\ \{S_1,\ S_2, ..., S_{l(s)}\}\ $ is the silhouette polygon, and $L(S),\ A(S)$ are the length and enclosed area to obtain feature vector presented in eq. \ref{eq31}. 
\begin{equation}\label{eq31}
f_1\ =\ (s_1,\ s_2, ..., s_{l(s)}, L(S), A(S))
\end{equation}
The $C$ is calculated as in subsection~\ref{uhl-para}. The \emph{shape of the foot}\cite{Naka00} is a feature based on 15 local foot widths and positions, while \emph{Eigenfeet}\cite{Turk91} is the projection of sub-sampled footprint onto feature space spanned by 20 principal components, both shape, and  Eigenfeet features can classify based on \emph{Manhattan distance}\cite{Craw2010}. The feature vector shown in eq.~\ref{eq32}: 
\begin{equation}\label{eq32}
f_2=(\omega_1, \omega_2,..,\omega_L)
\end{equation}
consists of exactly $L$ components given by eq.~\ref{eq33} for projection onto eigenspace.
\begin{equation}\label{eq33}
\omega_i=u^T_i \Phi
\end{equation}
Here $\Phi$ is given by eq.~\ref{eq34}
\begin{equation}\label{eq34}
\Phi\ \sim\  \sum^L_{i=1}\omega_i \ u_i
\end{equation}
and $u_i$ be the set of Eigenfeet $\forall i\in {1,2,..,L}$. After normalizing foot, it is equally divided into $N$ vertical slices $V_0,..., V_{N-1}$. Let $\chi$ be the characteristic function and $S_y$ be the $y-$monotone polygon to compute the width using eq.~\ref{eq35} of a segment of the foot of size $m\times n$ with $V_i \cap S_y, \ \forall i=0,1,.., N-1$.
\begin{equation}\label{eq35}
w_i=\frac{N}{n} \sum^n_{j=1}\ \sum^m_{k=1}\ \chi\ V_i \cap S_y\ (j,k)
\end{equation}
Then the feature vector for width is given as eq.~\ref{eq36} with $N=15$. 
\begin{equation}\label{eq36}
f_3\ =\ (w_0,..., w_{N-1})
\end{equation}
A \emph{Euclidian distance} based classifier classifies \emph{sole print} \cite{Kumar2003} which is a variance of 288 overlapping blocks in the edge-detected image; it also classifies \emph{toe length} which includes five toe lengths and four inter-toe angles. If $\alpha$ is inter-toe angle, then feature vector for toe length can be given by eq.~\ref{eq37}.
\begin{equation}\label{eq37}
f_4 = (L_1,\alpha_1, L_2, \alpha_2, L_3, \alpha_3, L_4, \alpha_4, L_5)
\end{equation}
\begin{equation}\label{eq38}
f_5=(\sigma^2_1, \sigma^2_2 ,.., \sigma^2_{288})
\end{equation}
The feature vector presented in eq.~\ref{eq38} defines sole print, which consists of extraction of the variance of 288 overlying chunks of size $24 \times 24 $ pixels each.  NIST's \texttt{bozorth} algorithm can classify \emph{minutiae} using the \texttt{mindtct} minutiae extractor on ballprint region following the big toe matcher\cite{Ko16}. For each value of $i$ from 1 to $n$, $f_i$ represents the elements of $f$, and for each value of $j$ from 1 to $m$, $t_j$ denotes the reference template for matching. If $e$ in eq.~\ref{eq39} is an evaluation function of Silhouette algorithm\cite{Turk91} then:
\begin{equation}\label{eq39}
e(f,t)=\alpha \ D(f,t) + \beta\ \mid f_{n-1}-t_{m-1}\mid + \gamma \mid f_n-t_m\mid
\end{equation}
where $D$ resembles silhouette data sets employing a cost function shown in eq. \ref{eq40}: 
\begin{equation}\label{eq40}
c(i,j)=(f_i-t_j)^2
\end{equation}
on DTW, and the minimum distance both the dataset is given by eq.~\ref{eq41}.
\begin{equation}\label{eq41}
D(f,t)=d(n,m)
\end{equation}
If the value of $i=j=1$, then $d(i,j)=0$, for a value of $i>1;j=1,\ d(i,j)=c(i,j)+d(i-1,1)$, similarly for $i=1;j>1,\ d(i,j)=c(i,j)+d(1,j-1)$, or otherwise $d(i,j)=c(i,j)+min(d(i-1,j),d(i,j-1),d(i-1,j-1))$\cite{Ross03}.

\subsection{Estimation of Stature from Footprint and Foot Outline Dimensions}
Ref.~\cite{Krishan08} had applied 2080 footprint and foot outline dimensions collected from 1040 adult male of 18 to 30 years old for ten and eight measurements respectively, for a particular tribe of criminal background which helps in an estimation of the physical build of a criminal to understand the meaning of T-1, T-2,...,T-5). His study also signifies the utmost importance of footprint into understanding the crime scene investigations. This study indicates statistically significant bilateral asymmetry of T-2 and T-5 length in the footprint and T-1 length, T-4 length, and breadth at the ball in foot outline. These features indicate a close relationship between the stature and these measures with the highest positive correlation coefficients of toe length measurements (0.82–0.87). Regression analysis gives better reliability with smaller mean errors (2.12–3.92 cm) in the estimation of stature (3.29–4.66 cm). This study motivates to conduct similar kind of research to carry out on other endogamous groups for proving the effect of genetics and environment in forensic terms from different parts of the world. The footprint always has a significant role in crime scene investigation.  

Later, Ref.~\cite{Moorthy2015} uses a sample of 400 adult Malay participants consisting of 200 males and 200 females using an inkless shoe print kit for investigation of features of the toes, elevations in the toe line, phalange marks, flatfoot condition, pits, cracks, corns and so forth. The occurrence of the fibularis-type foot is the largest, followed by tibialis-type. Then intermediate-type and medullaris-type are determined to be the least frequent in both the sexes, which is different sequence observed in north Indian population.
 
\subsection{Load Distribution Sensor using Fuzzy Logic}
\label{takeda09}
Ref.~\cite{Take09} has collected one step walking foot pressure images from 30 volunteers from 20 to 85 years old using a mat type load distribution sensor of 330mm $\times $1760mm size comprised of 128 vertical and 64 horizontal electrode sheets with a vertical interval of 7mm and a horizontal interval of 5mm. The fuzzy classifier is defined by eq.~\ref{eq42}.
\begin{equation}\label{eq42}
\mu_{person}\ =\ \sum^{38}_{i=0}\ \mu_{F,i}\ (f_ i(X))
\end{equation}
where $\mu_{person}$ is the fuzzy degree of a person, $X$ is a walk data, and $\mu_{F,i}$ is the total fuzzy degree, has applied to all the obtained dynamic feature weight movement and footprint feature foot shape during walking. The fuzzy classification of a person has inferred from out of six samples per volunteer five samples used for training and one sample for test data. This method gives 6.1\% EER and 13.9\% FRR  in verification ($1:1$ collation) and identification ($1: N$ collation).  
\subsection{Foot Pressure Change using Neural Network}
Ref.~\cite{Ye09} obtain the centre of pressure (COP) features (position and the movement) for a personal identification system captured from 11 volunteers (10 samples/volunteer) standing statically on the load distribution sensor with a slipper. The load distribution sensor Nippon Ceramic Co., Ltd., NCK is composed of 64 horizontal $\times$ 128 vertical electrodes sheets. The size of each such sheet is 330mm $\times$ 640mm, and the interval between two layers is 5mm. The inclination angle of the slipper is \SI{9.5}{\degree}. This system has extracted 16 features. Out of 10 samples of the dataset, five rows have used as the test set, and the remaining five has used for comparison. This system has developed with a k-out-of-n system and a neural network model under less information, time and small space and experimented input test data to both the plans. The correlation coefficients of the {\bf k-out-of-n} method for matching has calculated as eq.~\ref{eq43}.
\begin{equation}\label{eq43}
r=\frac{\sum^n_{i=1}(x_i-\overline{x})\ (y_i-\overline{y})}{\sqrt{\sum^n_{i=1}(x_i-\overline{x})^2\ (y_i-\overline{y})^2}}
\end{equation}
with $x_i$  be the test data in $t_i$, $y_i$ denotes the template in $t_i$, $\overline{x}$ and $\overline{y}$ express the average of the test and template data respectively.  The {\bf Neural Networks} based technique for recognition consist of 16 input neurons (number of features), 17 hidden neurons, and 11 output neurons (number of users). These methods show FRR of 28.6\% and FAR of 1.3\% for the k-out-of-n system and FRR of 12.0\% FAR of 1.0\% in neural network.
\subsection{Wavelet and Fuzzy Neural Network}
Ref.~\cite{Wang09}, gives a method based on Wavelet and Fuzzy Neural Network with a claimed recognition rate of 92.8\%.  Wavelet distinguishes the edge of the footprint to identify the statistical features of the four different shapes (triangle, ellipse, circular and irregular) of toe images such as angle, length, and area.  The feature vector given in eq.~\ref{eq44}: 
\begin{equation}\label{eq44}
\begin{split}
V=(v_1=E, v_2=r_c-r_i/r_i, v_3=S, v_4=AC, \\
v_5=\alpha_1, v_6=\alpha_2)
\end{split}
\end{equation}
is constituted by six parameters, where $E=p/q$ is the eccentricity of toe region, $v_2$ calculates the ratio of the radius difference between circumscribed circle and inscribe circle to the radius of the inscribed circle, $S=r_i/r_c$ is the sphericity of toe region, $AC$ calculates the area of toe region, $v_5$, and $v_6$ represents the angular parameters of the toe. This technique uses six-six membership functions of ellipsoidal toe shape, triangular toe shape, circular toe shape, and irregular toe shape. The sole judgment factor obtained from these membership functions is operated by AND to get the comprehensive judgment vector. The neural network has four input, and four output neurons for each shape types. Comprehensive judgment vector is compared with four model vectors to determine the distance vector to feed into neural networks to decide. A sum of 80 samples extracts 320 toe images for different shapes. 

\subsection{Fuzzy Artificial Immune System}
Ref.~\cite{Take10,Takeda2011} proposed a biometric authentication system based on One Step Foot Pressure Distribution Change and acquired footprints from the mat type load distribution sensor to get twelve features based on the shape, and twenty-seven features based on the movement of weight while walking from 10 volunteers and took the step data five times. The system setup is the same as one given in subsection~\ref{takeda09}. If $Y$ be the registered person set \emph{see ~\ref{eq45}},
\begin{equation}\label{eq45}
Y = \{y_1, y_2,..., y_n\}
\end{equation}
$M$ be the number of learning data except for $y_r$, $N$ be the number of learning data with respect to $y_r$, then for all $y_r$ for walking person $y_s$ calculates $\mu^{y_r}_{step}(y_s)$. Fuzzy if-then-else rules have created with fuzzy degree $P_i$ of legal person denoted by $f_i(X)$, and $s_i$ be the standard value of legitimate user by training data. The first rule says: \\ \\
{\scriptsize {\texttt{{\bf{RULE 1}}: IF feature $f_i(X)$ is CLOSE to the standard $s_i$, THEN the degree $P_i$ is high.} \\}}\\ 
The fuzzy classifier has developed for each feature which is trained by a clonal selection algorithm in the Artificial Immune System(AIS). After clonal selection phase, personal authentication for a step data phase utilizes the second rule, which is:\\ \\
{\scriptsize {\texttt{{\bf{RULE 2}}: IF classification performance $Score_i$ of training is HIGH, THEN the degree $W_i$ is high.}}}\\ \\
where $W_i$ be the weight of the feature.  Final phase is the authentication to authenticate a legal person with fuzzy degree $\mu^{y_r}_{Step} (y_s)$ by the step data.  AIS has evaluated by the five-fold cross validation technique gives FRR of 4.0\%, FAR of 0.44\% and EER of 3.8\%. 
\subsection{Modified Sequential Haar Energy Transform (MSHET)}
Ref.~\cite{Kumar10} introduce modified sequential haar energy transform technique. The human footprint is robust as it does not change much over the time, universal, easy to capture as it does not need specialized acquisition hardware. Initially, the algorithm starts with acquiring the left leg images taken in different angles without provisioning any special lighting in this setup from 400 subjects. The obtained images are then normalized for the key points and cropped respectively and converted to grayscale from RGB. Feature extraction methods Linear Predictive Coding (LPC), and Linear Predictive Cepstral Coefficients (LPCC) given by \cite{Hariharan2012} has applied to the obtained images to enhanced it further by transforming to a matrix format separated into odd and even matrices of 256 $\times$ 256 pixels. Sequential Modified Haar(SMH) transforms employed to the resized images ($4\times4$ blocks) to retrieve a Modified Haar Energy (MHE) features determined by eq.~\ref{eq46}: 
\begin{equation}\label{eq46}
MHE_{i,j,k}=\sum^4_{p=1} \sum^4_{q=1} (C_{p,q})^2
\end{equation}
for the level of decomposition $i$, neighborhood detail $j$ is, and block number $k$ from 1 to 16. The SMH wavelet can map integer-valued signals onto integer-valued signals yielding the property of comprehensive restoration. The Euclidean distance compares MHE feature with the feature vectors stored in the database and yields 92.375\% of accuracy. Talukder and Harada~\cite{Talukder07} provided the methodology to use Haar transform in images.  
\subsection{Multibiometrics Rank-Level Fusion}
Ref.~\cite{KumarS10}reviews an approach for the personal recognition employing a nonlinear rank-level fusion of multiple biometrics representations. This approach gives a direction to apply rank-level fusion in the footprint biometrics. It compares results from openly available multi-biometrics (NIST BSSR1 database from 517 subjects) scores and real palm print biometrics data to determine the rank-level fusion employing Borda count, logistic regression / weighted Borda count, highest rank system, and Bucklin system. The empirical results presented suggest that significant performance improvement on the recognition accuracy from 94\% to 99\%. The precise experimental results also indicate that the nonlinear rank-level approach excels over other rank-level fusion methods.
\subsection{Dynamic Plantar Pressure}
Ref.~\cite{Pata11} has used a sum of 1040 dynamic foot pressure patterns which are usable immediately and collected from 10 volunteers (104 samples per subject) based on automatic spatial alignment for achieving high classification rate (CR) of 99.6 \% regardless of automatic dimensionality reduction. These patterns reveal the expeditions of all body parts are different leads to the conclusion that pointedly indicates inter-subject pressure pattern uniqueness. The pre-processing requirement is also minimal, and as they might collect while uninterrupted gait using in-floor systems, foot pressure-based identification resembles to have a broad perspective in the security and health industries.

\subsection{Discrete correlation of Footprint Image}
According to the survey presented by Ref.~\cite{Kumar11a, Kumar12c}, the last decade was devoted to footprint-based feature extraction for personal recognition. They have discussed and concluded some most prominent feature extraction techniques. The algorithms are commonly divided into three major stages (i) image pre-processing-is the method used to diminish the noise, (ii) feature extraction from images such as Center of Foot Pressure (COP), geometric constraint, toe shape, triple feature, vibration feature, tactile sensor feature, image feature, walking pattern, stepping pattern, footprint arch, location and weight and (iii) template matching to improve the performance by enhancing the matching score. The authentication based on footprint owns sufficient security without any shortcoming of relying on sensitive data required by high-security applications.

\subsection{Orientation Feature--Newborn Baby}
Ref.~\cite{Jia2012} has developed an online footprint-based system for identification of newborn and infant, which is a critical issue for the places where multiple births occur. This approach captured high-quality images of size $691\times 518$ as compared to traditional offline methods. The offline footprint images go obscured due to: (i) Use of incompetent stuff like ink pads, paper, cylinder and so forth; (ii) Lack of training of personnel for footprint image acquisition; (ii) Oily matter on baby's skin gives non-recognisable prints; (iv) The epidermis is quickly distorting the ridges upon touch and filling the valleys within ridges with ink due to less thickness; (v) Tiny size of the infant's ridges. The preprocessing phase changes the orientation and scale to normalize the footprint image. Initially, they start with image segmentation to extract footprint \( G(x,y)\) with threshold \(T\) to get the segmented binary image. Next phase calculates the centre of footprint \(C(x,y)\) in binary image \(B(x,y)\) by using eq.~\ref{eq47}: 
\begin{equation}\label{eq47}
C(x,y)=centre\left\{ \frac{\sum_x x \sum_y B(x,y)}{\sum_{x,y} B(x,y)} ,\frac{\sum_y y \sum_x B(x,y)}{\sum_{x,y} B(x,y)} \right \} 
\end{equation}
The obtained image is then normalized to \(C(x,y)\) and a region of interest (ROI) is cropped. The database consists of 1938 images from 101 newborns applied in the recognition stage on four orientation feature-based methods, Ordinal Code, BOCV, Competitive Code, and Robust Line Orientation Code.  The highest recognition rate achieved has 96\%. 

\subsection{Manifold Feature Extraction using PCA and ICA}
Ref.~\cite{Kumar12j, Kumar12, Kumar13, Kumar13j} gives a method based on manifold feature extraction. Ref.~\cite{Siva15} has extracted fingerprint minutiae feature and PCA footprint features of a newborn on the Raspberry Pi system. Starting from capturing the footprint, preprocessing provides binary images; these binary images then yield COP data; subsequently, feature extraction phase extracts features; PCA trains the data, and stochastic gradient descent algorithms have utilized to match the footprint for recognition. PCA can do recognition with SVM, or Wavelet-based Fuzzy Neural Network can also be used for recognition. The feature has extracted based on Gabor filter and Discrete Wavelet Transforms (DWT). The maximum recognition rate obtained has 90.35\%. 

\begin{figure}[h]
\centering
\includegraphics[width=0.5\textwidth]{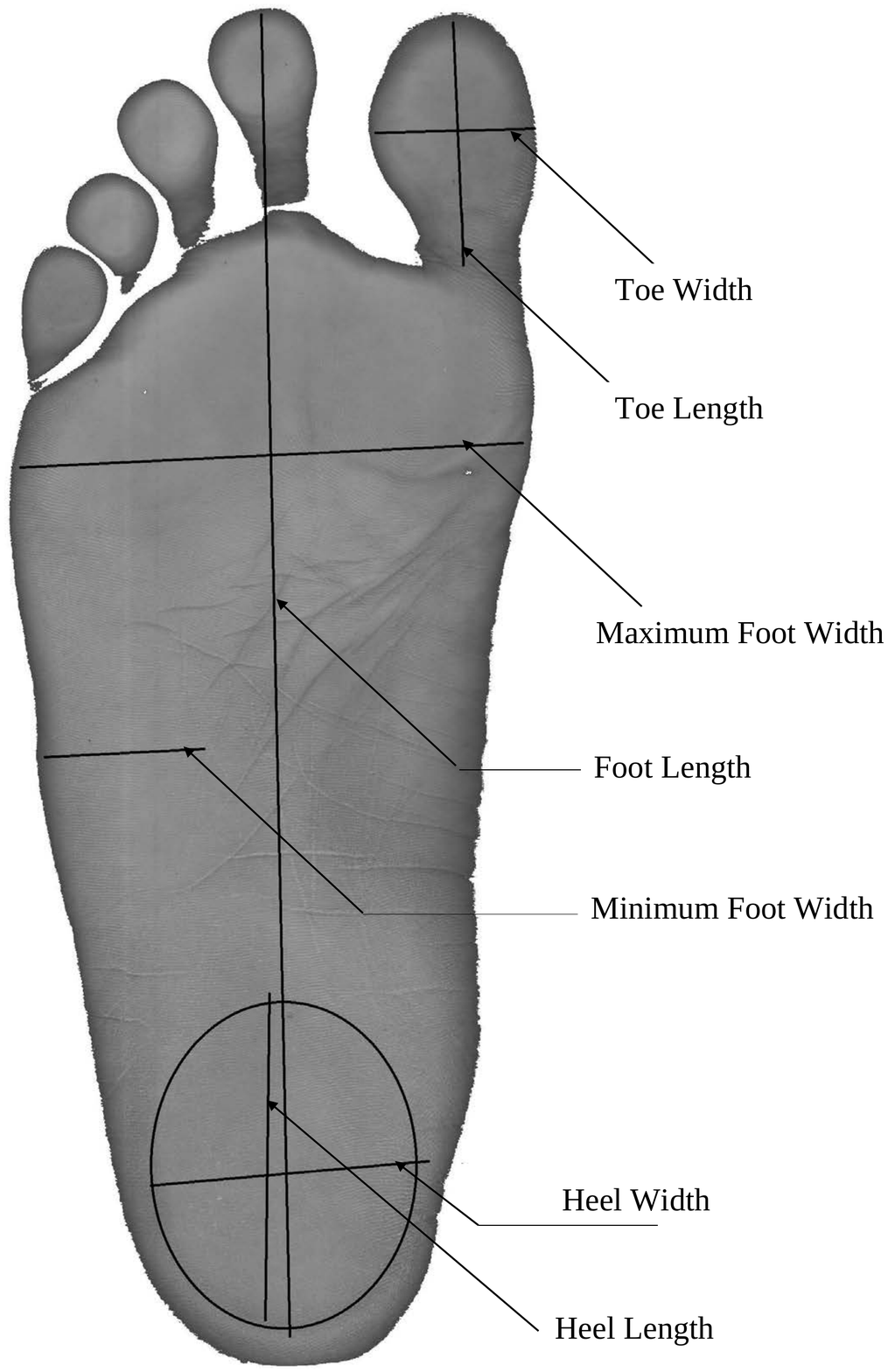} 
\caption{Different parts of the footprint to estimate statistical details.}\label{2foot}
\end{figure}

Ref.~\cite{Kumar15} has used multiple features to automated identification based on segmentation. Figure \ref{2foot} exhibits different parts of the footprint to estimate the statistical details. The system is further extended to identify the human by footprints using ICA and PCA. This system covers various mathematical computations of the footprint indices for the identification.  A hundred-time fraction of minimum foot width and overall foot width gives the Instep-Foot Index. Ball-Foot Index can have calculated as a hundred-time fraction of maximum foot width and overall foot width. A hundred-time portion of heel width and heel length yields Heel-Index. Toe-Index is calculated as a hundred-time fraction of the toe width and toe length. Likewise,  a hundred-time fraction of maximum foot width and heel width estimates Ball Heel Index. Finally,  a hundred-time portion of the toe width and maximum foot width determines the Toe Ball Index.

Ref.~\cite{Kho15} exhibit the use of PCA and Independent Component Analysis (ICA) for footprint recognition. PCA initially computes the covariance matrix using eq.~\ref{eq48}: 
\begin{equation}\label{eq48}
C=(X-\bar{x}) \times (X-\bar{x})^T
\end{equation}
$\forall$  mean \(\bar{x}\) in dataset \(X\). The relationship between eigenvectors \(V\)  and \(C\) in \( V^{-1} \times C \times V = D\) gives eigenvalues \(D\). \(V\)  is stored in  descending order of \(D\) to evaluate the projected data \(\wp\) in \(V\)-dimensional vector space by relation eq.~\ref{eq49}: 
\begin{equation}\label{eq49}
\wp=[V^T (X-\bar{x})^T ]^T 
\end{equation}
The computation in ICA begin with centering to get centered observation vector of \(x\) as shown in eq.~\ref{eq50}. 
\begin{equation}\label{eq50}
x_c = x-\bar{x}
\end{equation}

After getting  unmixing matrix using the centralized data the actual estimates of the independent components can be obtained by  eq.~\ref{eq51}. 
\begin{equation}\label{eq51}
\hat{s}=A^{-1} ( x_c+\bar{x})
\end{equation}
The observation vector \(x\) is now transformed to whitening vector \(xw\) and satisfies the covariance matrix \( E\{x_w x^T_x\} \) of \(x_w\)  to unity. 
\begin{equation}\label{eq52}
 E\{x x^T\} = V D V^T
\end{equation}
Eq.~\ref{eq52} presents the breakdown of the covariance matrix of \(x\). Further, eq.~\ref{eq53} gives whitening transform of \(x_w\). 
\begin{equation}\label{eq53}
x_w=V D^{-1/2} V^T x
\end{equation}
Also, Whitening converts the mixing matrix into a new orthogonal form by eq.~\ref{eq54} 
\begin{equation}\label{eq54}
\begin{split}
x_w=V D^{-1/2} V^T A s= A_w s\\
hence, \  E\{x x^T\} = A_w E\{s s^T\}A^T_w =I
\end{split}
\end{equation}
Therefore, whitening reduces the number of components to be estimated. For 21 subjects \cite{Kho15} claims ICA with the RR of 97.23\% and PCA 95.24\% RR.

\subsection{Foot-boundary for Crime Scene Investigation}
Saparudin and Sulong~\cite{Saparudin16} elaborated a method based on global features of fingerprints for automatic classification. Later, Ref.~\cite{Das16} gives computational analytics of crime scene investigation, which is a meeting point of science, logic, and law using foot boundary detection. Mostly, in the manual crime scene, the investigator investigates based on blood spots, hair, fingerprints, and available piece of belongings makes the task very long and slow. Foot biometry has mostly ignored so far. This approach takes the high-definition pictures of the available footprint of any form. These images then used for identification purpose based on foot boundary.

\begin{table}[!t]
\caption{Comparison of Litrature (NA-Not Applicable)\label{comparison}}
{
{\scriptsize{
\begin{tabular*}{20pc}{@{\extracolsep{\fill}}lcr@{}}\toprule
Main  & Number of  & Recognition \\
Idea [Author]  &Subjects &  Rate \\
\midrule
Euclidean distance \cite{Naka00}  & 10    & 85\% \vspace{1mm} \\
Viewpoint based  
& 25 
& 76\% to 100\%.		\\
template matching \cite{Collins02}
&
& \vspace{1mm} \\
HMM blended 
& 11
& 64\% to 80\%.                            \\
Levenberg-Marquart 
&
&
\\
learning\cite{Jung02, Jung03a, Jung03b}
&
&
\vspace{1mm} \\
Invariant gait  
& 12 
&  Not stated \\
recognition \cite{Kale03}
&
&
\vspace{1mm} \\

Gauss curvature\cite{Wang04}
& 70
& not stated              \vspace{1mm}  \\
HMM applied
& 71
& 86\% \\
in gait 
&
&
\\
stances\cite{Boul05}
&
&
\vspace{1mm} \\ 
View transformation \cite{Makihara2006} 
& 20 
& VR 10\%
\vspace{1mm} \\
Patient's 	
& not for  
& NA \\
walking 
& personal
&  \\
pattern\cite{Kura05}
& identification
& \vspace{1mm} \\
PARAchute project \cite{Hew07}
& NA
& not stated
\vspace{1mm} \\
ANN Dynamic                                   
& 10
& 92\%                             \\
Footprint\cite{Yun07}
&
& to 96 \% \vspace{1mm}  \\
Eigenfeet, Ballprint 
& 16                                                                                                                                                                                            & 97\%                             \\
and Foot geometry \cite{Uhl07,Uhl08, Wild08}
&
&
\vspace{1mm} \\
Bilateral asymmetry\cite{Krishan08} 
& 1040 
& NA 
\vspace{1mm} \\
High-resolution 
& 11 
& 92.3\% 
\\
pressure sensing\cite{Qian2008,Qian2010}
&
&
\vspace{1mm} \\
wavelet and fuzzy                                  
& 80
& 92.8\%                             \\
neural networks\cite{Wang09}
&
&
\vspace{1mm} \\
Fuzzy Gait\cite{Take09}                              
& 30
& 86\%            \vspace{1mm}  \\
ANN \& K-out-of-n
& 11 
& FRR of 28.6\% to 1\%
\\
COP features \cite{Ye09} 
&
&
\vspace{1mm} \\
nonlinear 
& 517 
& 94\% to 99\%
\\
rank-level fusion\cite{KumarS10}
&
&
\vspace{1mm} \\
One Step Foot     
& 10
& FRR 4.0\%            \\
Pressure AIS \cite{Take10, Takeda2011}
&
&
\vspace{1mm} \\
MHE Euclidean  
& 400
& 92.375\%
\\
distance \cite{Kumar10}
&
&
\vspace{1mm} \\
Radar Doppler 
& 5
& 91.2 \% \\ 
gait image-MLP\cite{Tiv10}
&
&
\vspace{1mm} \\
Part-based  
& NA
& EER 0 to 0.5\%  
\\ 
clothing gait\cite{Hossain10}
&
&
\vspace{1mm} \\
Soft biometric gait\cite{Ioannidis07, Moustakas10}
& NA
& FRR 4\% to 15\%
\vspace{1mm} \\ 
UDSA view-invariant
& 124
& 50\% to 90\% 
\\
gait signal \cite{Lu2010}
&
&
\vspace{1mm} \\ 
Foot pressure \cite{Pata11} 
& 10
& 99.6 \% 
\vspace{1mm} \\ 
Discrete correlation  & NA. 
& Not Stated               \\
PCA \cite{Kumar11a, Kumar12, Kumar13, Kumar12c}                                                                                                                                                          
&
&
\vspace{1mm} \\
Online system 
& 101 
& 96\%
\\
(newborn-infant) \cite{Jia2012}
&
&
\vspace{1mm} \\
Wavelet based Fuzzy 
& NA
& 90.35\%. 
\\
NN-PCA \cite{Kumar12j, Kumar12, Kumar13, Kumar13j}
&
&
\vspace{1mm} \\
Ink less shoe
& 400 Malay  
& not stated
\\
 print kit \cite{Moorthy2015}
 &
 &
\vspace{1mm}  \\
ICA and PCA \cite{Kumar15}
& NA
& not stated
\vspace{1mm} \\
PCA and ICA \cite{Kho15}
& 21                                                                                                                                                                                              & 95\%                             
\vspace{1mm} \\
Foot boundary detection \cite{Das16} 
& NA
& not stated
\vspace{1mm} \\
ANN \cite{Hash16} & 40  & 92\%\\                                                                                                                                                                                                                                                           

\hline
\end{tabular*}}}}
\end{table}
\subsection{ANN Foot Features}
Ref.~\cite{Hash16} have extracted sixteen geometric foot features of human footprints collected from 40 subjects. The ANN composed of a 16-node input layer, 11-node hidden layer, and 6-node output later to recognize 40 items for testing : training pair of 40:120 images. They have used ANN for the feature extraction and recognition with the highest recognition rate of 92\%.

\section{Results and Discussion}

Ref.~\cite{Rob78} had analyzed the uniqueness of human footprints. Footprint (and foot) morphology is significant since it explains the unique character of each person's footprints\cite{Naka00}. Hospitals used footprints to identify an infant by its footprints while Austria utilized its legal capabilities \cite{fernadez1953classification,Kapil12}. Since footprints are not intended to support large-scale high-security applications, such as electronic banking, the storage of features does not necessarily imply security threats. A person without having hands can use footprint biometrics in the legal capacity for identification. On the contrary, due to the practice of wearing shoes, it is difficult for imposters to obtain footprints for forgery attacks. Thus, footprint-based recognition may correspondingly be the best substitute for highest-security applications this motivates to develop an access control system based on footprint expected to have the highest recognition rate without the need for additional hardware.

\begin{figure*}[!h]
\includegraphics[width=15cm]{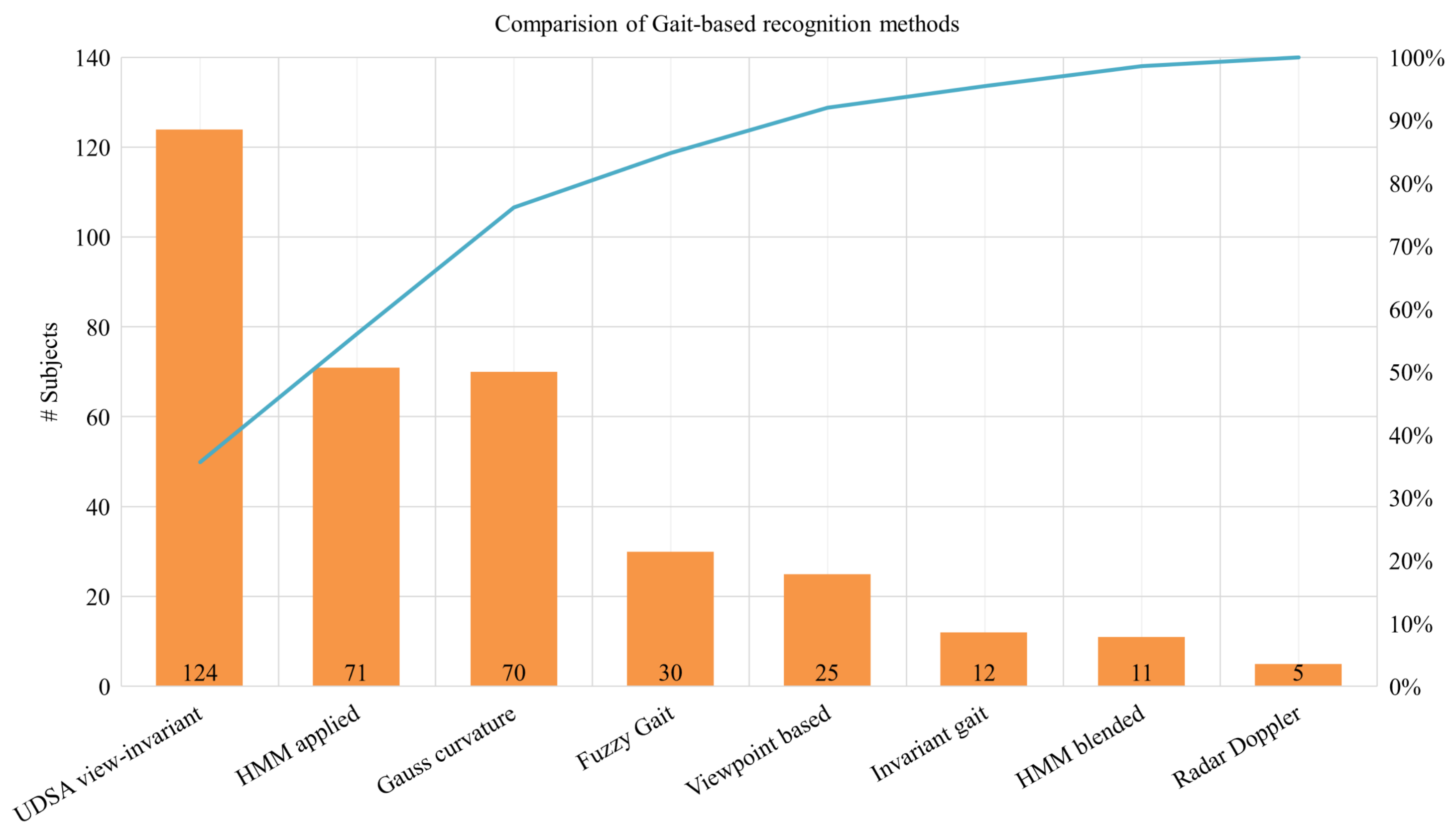}
\centering
\caption{Comparison of gait-based methods of personal identification.}
\label{fig-gait}
\end{figure*}

\begin{figure*}[!h]
\includegraphics[width=15cm]{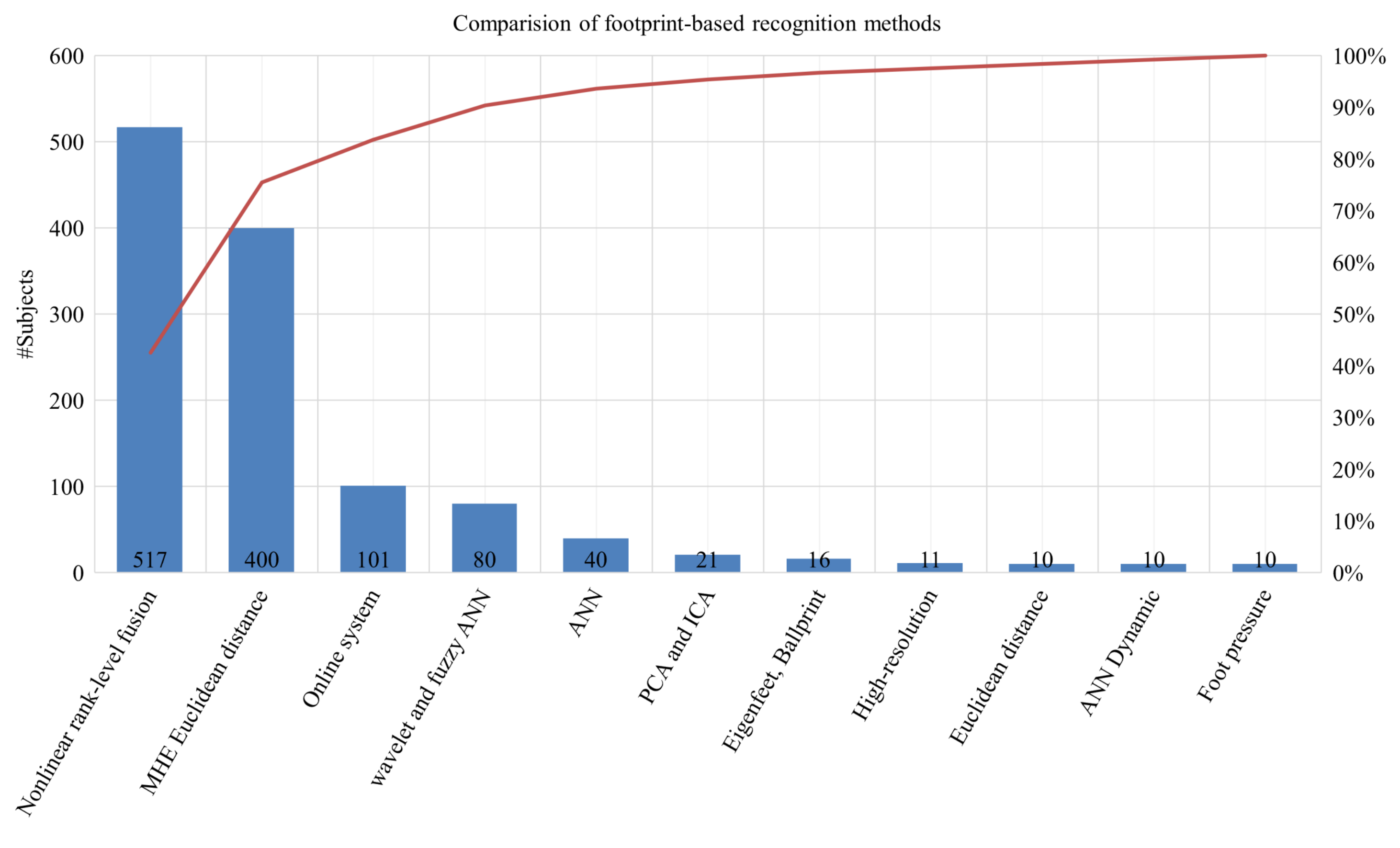}
\centering
\caption{Comparison of footprint-based methods of personal identification.}
\label{fig-foot}
\end{figure*}
\begin{figure*}[!htbp] 
    \centering
    \begin{subfigure}[t]{0.5\textwidth}
        \centering
        \includegraphics[height=2.2in]{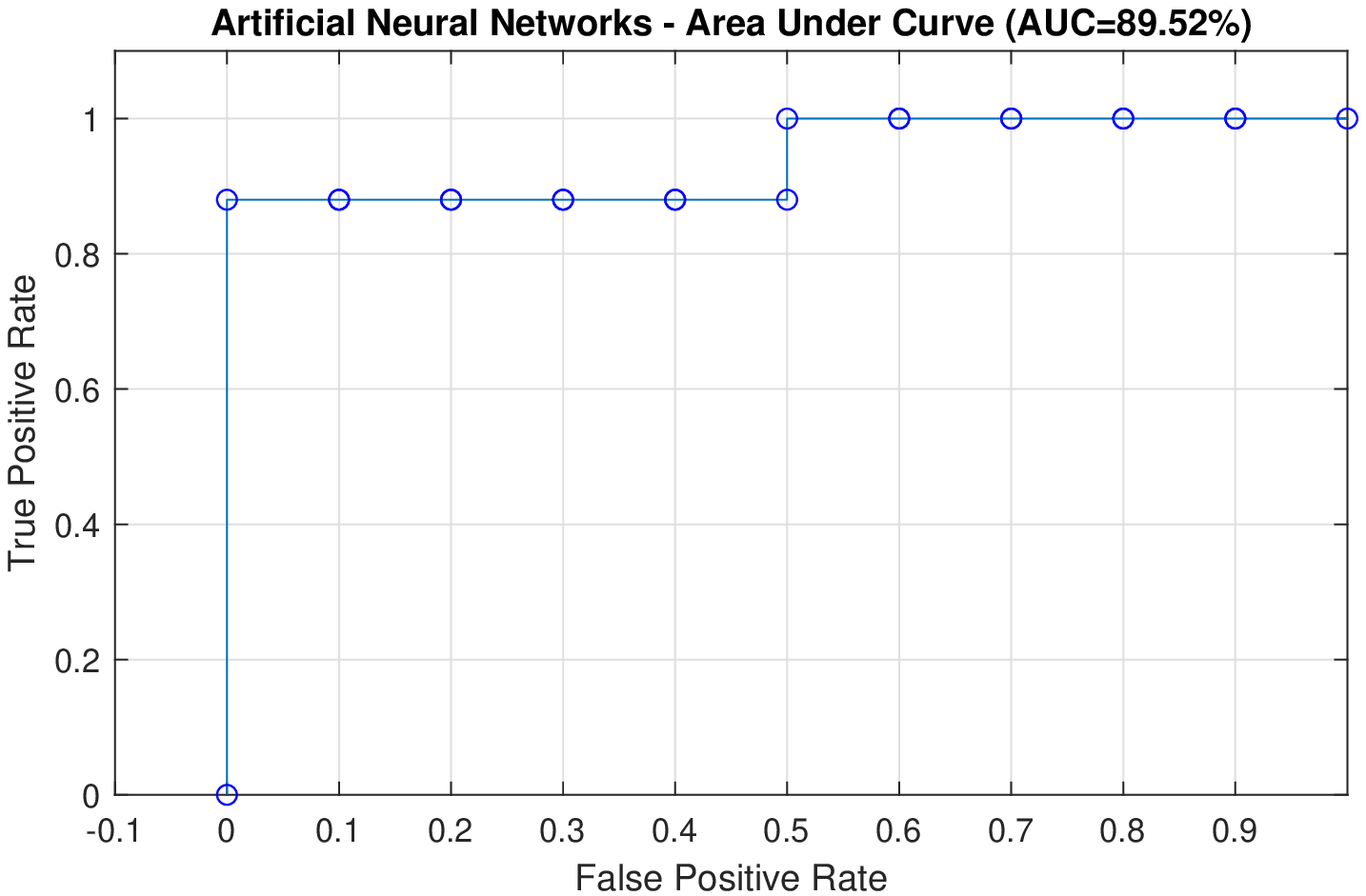}
        \caption{ANN}
    \end{subfigure}%
    ~ 
    \begin{subfigure}[t]{0.5\textwidth}
        \centering
        \includegraphics[height=2.2in]{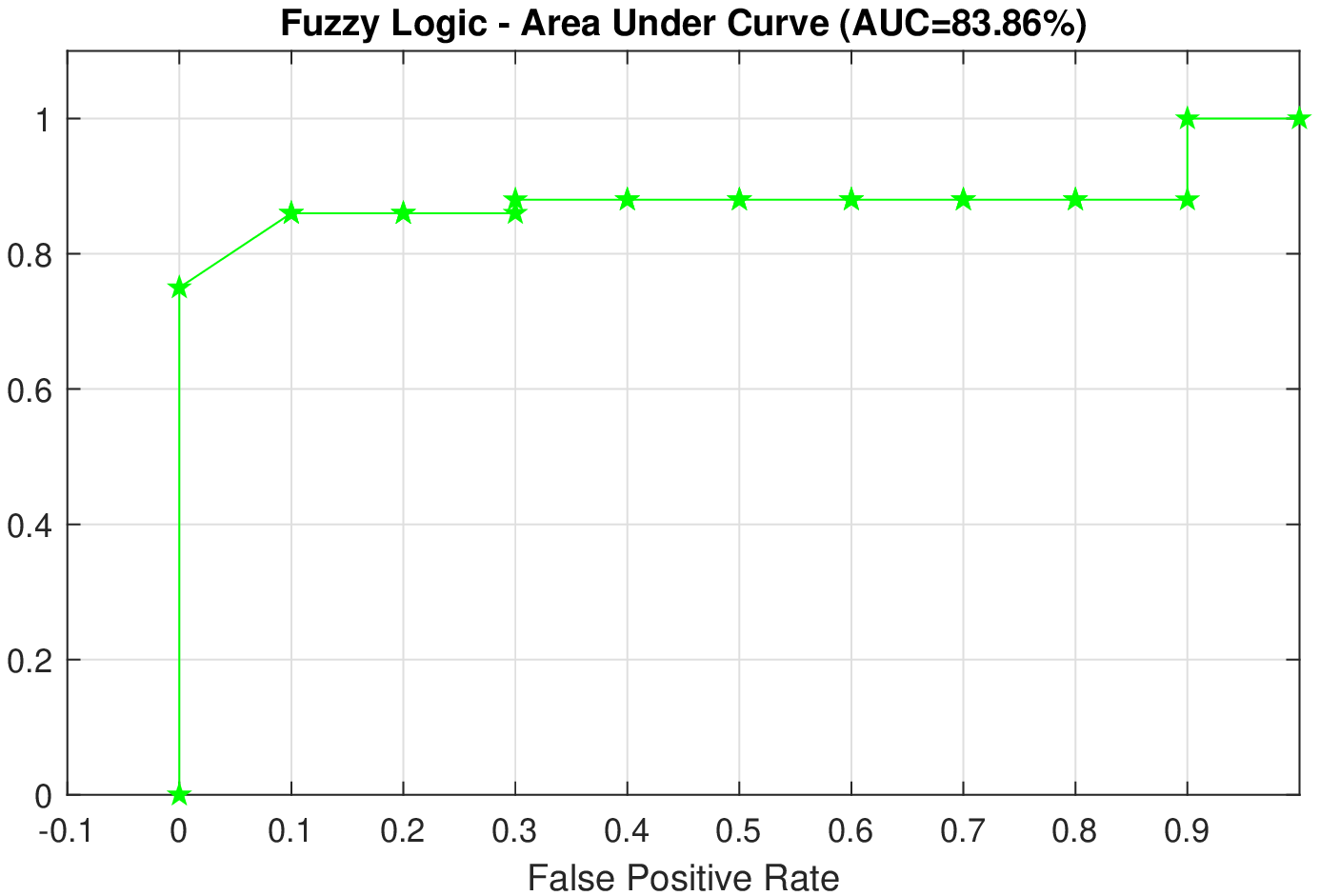}
        \caption{Fuzzy Logic}
    \end{subfigure}
    ~
    \begin{subfigure}[t]{0.5\textwidth}
        \centering
        \includegraphics[height=2.2in]{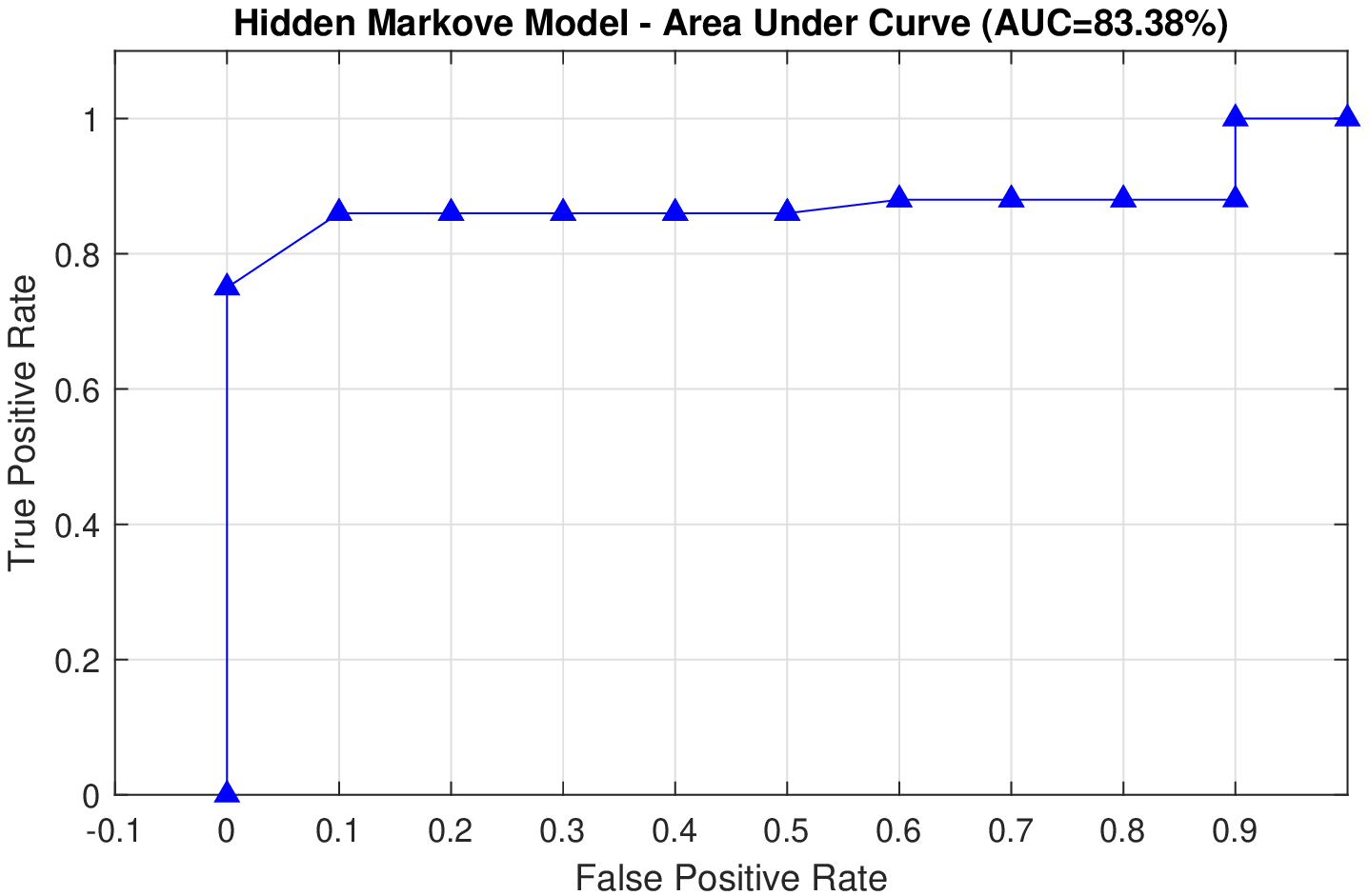}
        \caption{Hidden Markov Model }
    \end{subfigure}%
   	~
	\begin{subfigure}[t]{0.5\textwidth}
        \centering
        \includegraphics[height=2.2in]{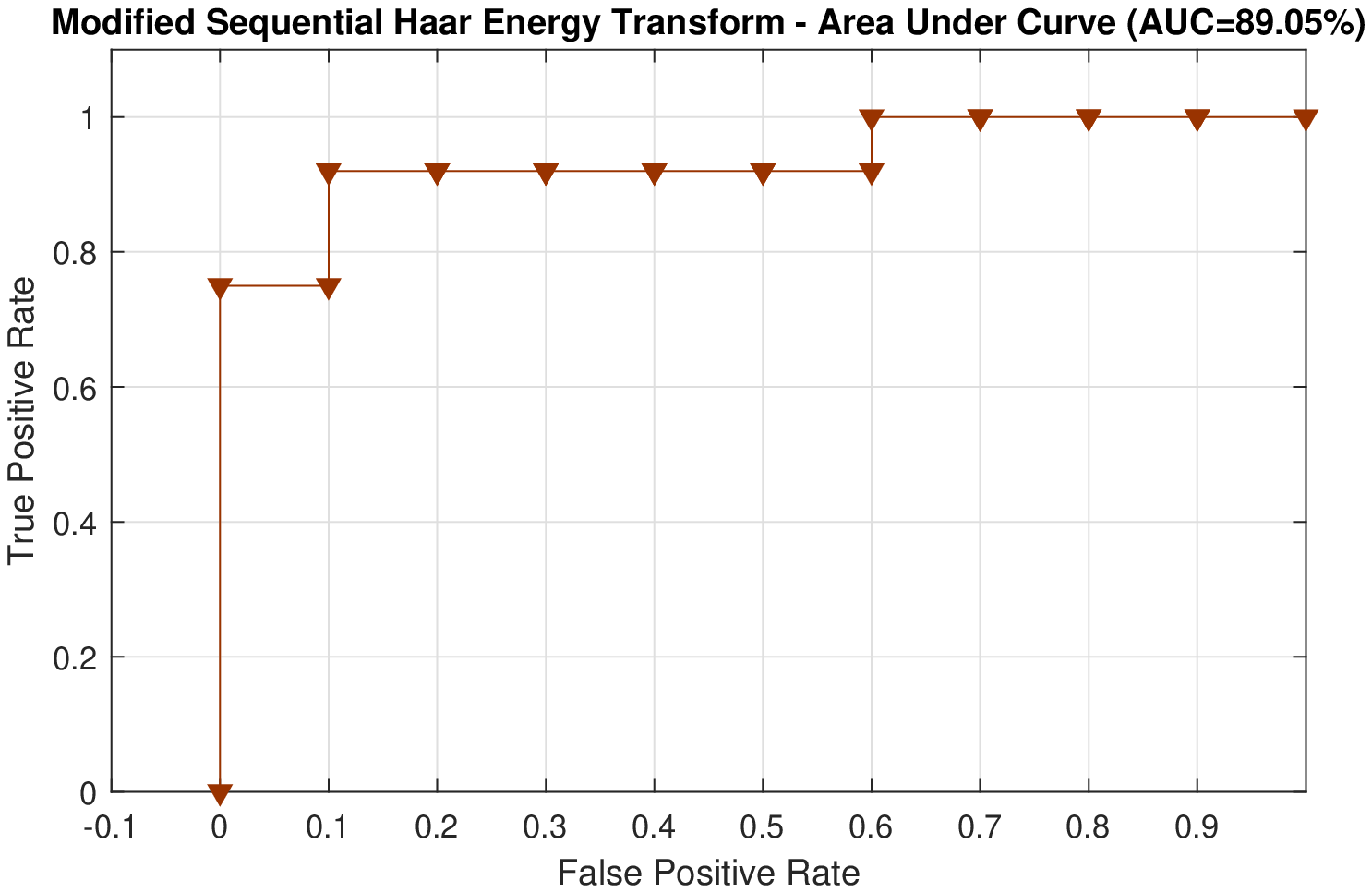}
        \caption{MSHET}
    \end{subfigure}
	\begin{subfigure}[t]{0.5\textwidth}
        \centering
        \includegraphics[height=2.2in]{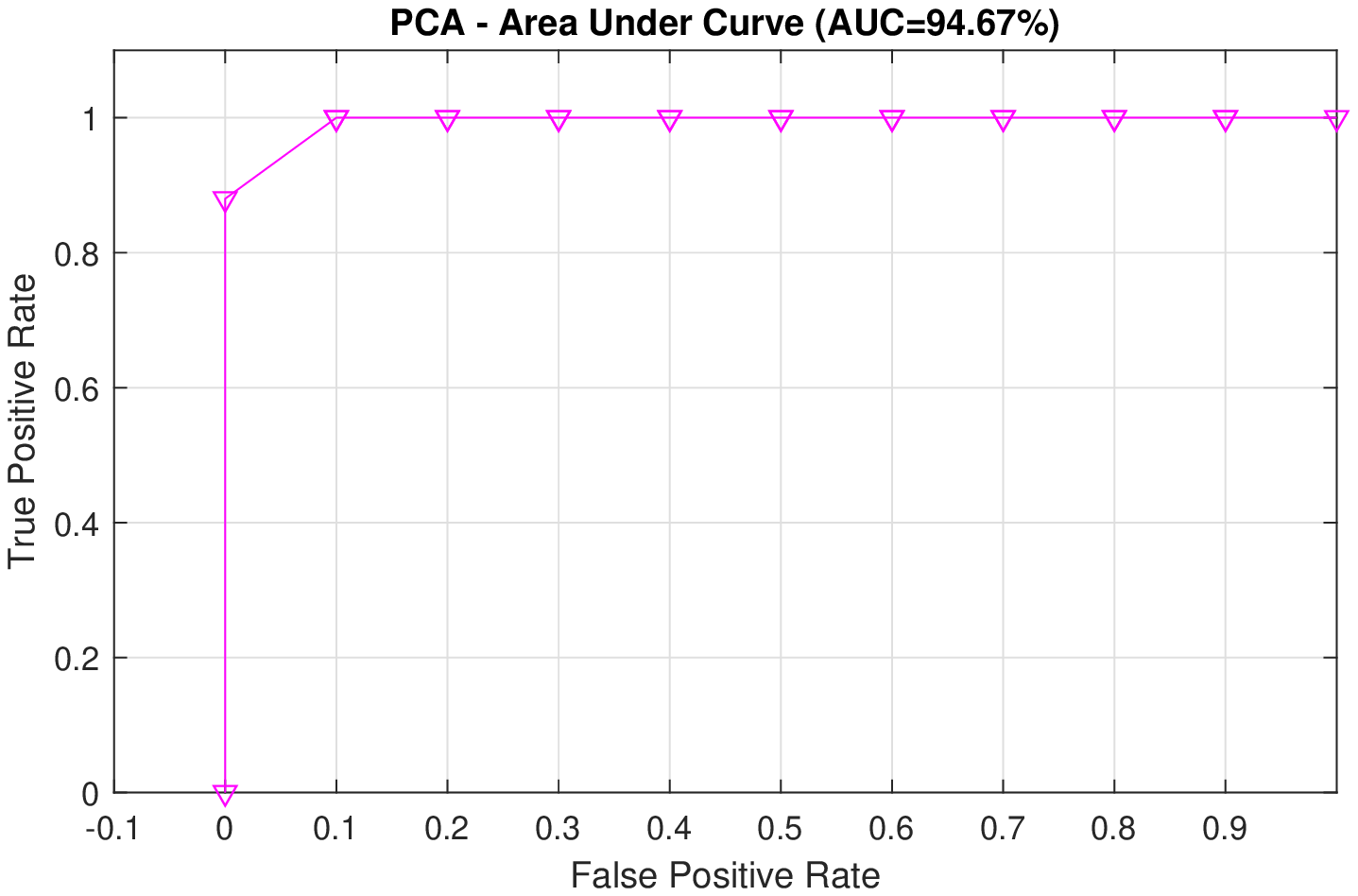}
        \caption{PCA}
    \end{subfigure}
    \caption{Comparison of performance of different algorithms for footprint-based biometric identification.} \label{fig9}
	\end{figure*}

Figure~\ref{fig-gait} gives a comparison of eight gait-based personal identification methods.  The x-axis represents the methodology used, left y-axis gives the number of subjects involved, and the right y-axis represents the recognition rate of the particular method. The blue line in this figure characterizes that as there is an increase in the number of subjects correspondingly the recognition rates decreases.Figure~\ref{fig-foot} gives a comparison of eleven footprint-based personal identification methods.  The red line in this figure characterizes that as there is an increase in the number of subjects correspondingly the recognition rates decreases.

Table~\ref{comparison} covers latest review summary for diffrent footprint-based biometric identification. Ref.~\cite{Jung02,Jung03a,Jung03b,Jung04} developed a technique to calculate  COP(Centre of Pressure) based on Hidden Markov Model. Ref.~\cite{Kumar12}encompasses a shoe-type pressure sensor and HMM(Hidden Markov Model). A comprehensive evaluation model comes into existence for establishing to recognize the toe shape using the neural network and fuzzy logic by Ref.~\cite{Jang93}. Self-organizing map \cite{Kohonen2001} for automation of the process, ART2 \cite{Shin2007} for optimization of footprint recognition, a trace transform technique for parallel line \cite{Kadyrov2001}, UbiFloor2 \cite{Yun07} using neural network, principal component analysis (PCA) \cite{Kumar12}, wavelet transform \cite{Kumar10,Li02} based footprint recognition; these are some methods applied during the last decade. Multimodal biometric is another method of person identification\cite{Gonzalez06,Al2013,Uhl07}. \cite{Wayman1996}  introduces a different classification of biometric. According to \cite{Wild08}, identify target application domains for footprint-based authentication. Ref.~\cite{Kho15,Kho16} put forward texture and shape oriented footprint features using PCA and Independent Component Analysis (ICA). Artificial Neural Network (ANN) based work for footprint feature extraction, and pattern recognition has been carried out to get a recognition rate of 92.5\% \cite{Hash16}.  

Figure~\ref{fig9} compares AUC (Area Under the Curve) performances for footprint biometric-based personal identification. AUC can be calculated by $\int_0^1 TPR(x)dx$, where $x \sim U[0,1]$ is the FPR of the footprint classification. The ANN-based approach gives 89.52 \% of AUC. The Fuzzy Logic based technique reported 83.86\% of AUC parameter while Hidden Markov Model give 83.38 \% AUC, on the other hand Modified Sequential Haar Energy Transform method said 89.05 \% AUC performance and finally, the PCA based method report 94.67 \% AUC, which is the best performance among all these five approaches.
\subsection*{Major Findings}
\begin{itemize}
\item Human gait gives good performance with less number of subjects, as the number of subjects increases it degrades the identification performance;
\item Human gait is suitable for health-care and other allied domain;
\item Human gait is also useful for thought recognition of a person;
\item Human footprint-based identification is a better candidate against other biometric traits;
\item The person without hand can give his/her footprints for biometric-based identification;
\item Comparatively increase in the number of subjects in the footprint-based approach gives a significant result hence this method proves its usefulness in personal identification;
\item Based on above finding human footprints are highly recommended for implementation of footprint-based biometric system in legal capacities; and 
\item The footprint and gait based biometric system proves its applicability in cyber-forensic examination.
\end{itemize}

\section{Conclusion and Future Scope}
Two different way of the foot has discussed here. First is based on behaviour or gait, and the second is purely based on the human footprint. The gait-based method takes a large amount of data per person to recognize.  One can realize the thought whether it is positive or negative, of a person while walking. The gait significantly gives the status of a patient, whether recovering or not. It also helps in determining the crime scene and create plausible theories. The bulk of images slows down the processing capabilities. It is therefore not possible to keep gait record of every individual. Most of the researchers took gait as patient care method, not a personal recognition method. Through the comprehensive literature survey, the gait-based approach is not suitable for personal identification.

On the other hand, the footprint-based process requires very few samples per person. To identify a person, both type of features (morphology, statistical) helps. This method is applicable at high-security zones, such as the airport, crowded places prone to a terrorist attack, silicon chip manufacturing agency, nanotechnology research laboratories, public wellness zones. The footprint-based method also plays a significant role in crime scene identification and creation of theories. This technique requires the barefoot person to capture the footprint. With the bulk of data, the recognition rate is higher as similar to a fingerprint-based method. The present study also explained the legal capability of the human footprint. 

The author has published a multispectral human footprint dataset for the evaluation of different techniques. The readers can utilize this data for their research in testing of performances of their algorithms. One can also use this dataset with various tools such as AzureML, BigML, MATLAB, Weka \textit{etc.} for future use\cite{kapildata15}. 
\section*{Acknowledgments}\label{sec11}
The author would like thanks to Ph.D. Supervisor Dr. Sipi Dubey from Rungta College of Engineering and Technology, Bhilai, India, who provided insight and expertise that greatly assisted the research, although she may not agree with all the interpretations/conclusions from this paper.

\ifCLASSOPTIONcaptionsoff
  \newpage
\fi



%

\bibliography{sample}


%

\ifCLASSOPTIONcaptionsoff
  \newpage
\fi



%
\begin{biography}[{\includegraphics[width=1in,clip,keepaspectratio]{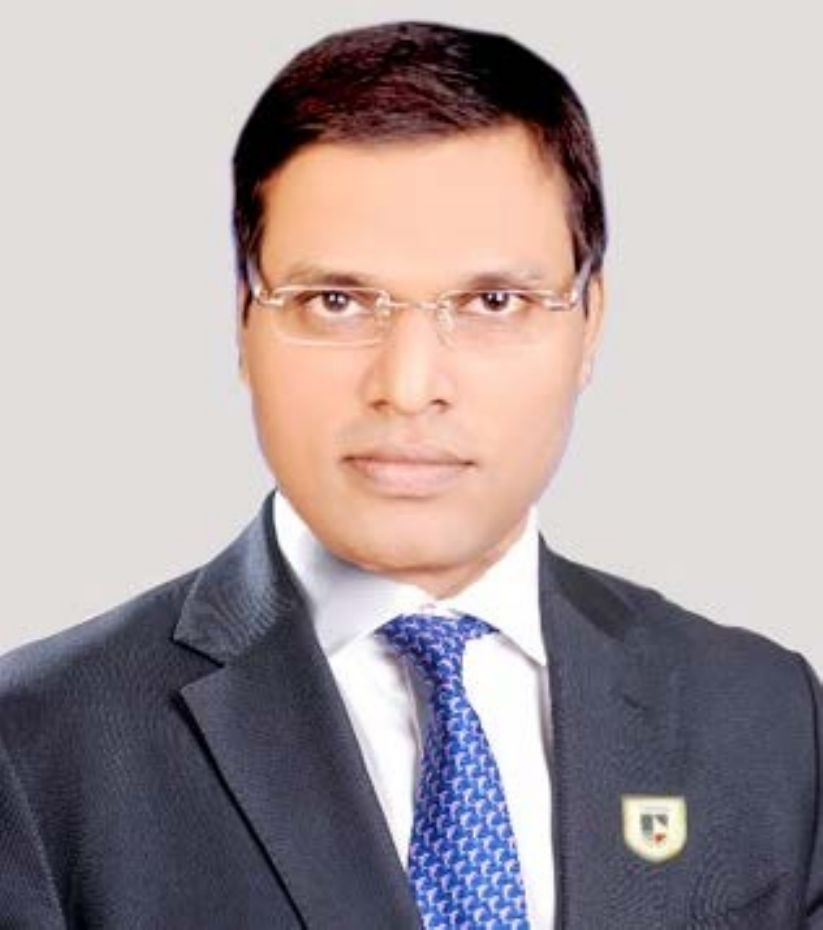}}]{ Dr. Kapil Kumar Nagwanshi} is working as an Assistant Professor at MPSTME Shirpur Campus, SVKM's Narsee Monjee Institute of Management Studies Mumbai, India.
He is a Ph.D. from Chhattisgarh Swami Vivekanand Technical University,
Bhilai. His research area includes biometrics, GPU computing,
neural networks, fuzzy logic and predictive analytics. He has published more
than 25 research papers in referred journals and conferences.
\end{biography}








\end{document}